\documentclass[10pt]{article}

\usepackage[no-math]{fontspec}
\usepackage[english]{babel}
\usepackage{csquotes}
\usepackage{lmodern}          % Latin Modern Roman (built-in)
\newfontfamily\arabicfont[
  Script=Arabic,
  Scale=1.1,
  BoldFont=Amiri-Bold.ttf,
  ItalicFont=Amiri-Italic.ttf,
  BoldItalicFont=Amiri-BoldItalic.ttf
]{Amiri-Regular.ttf}

\usepackage[nopatch=footnote]{microtype}
\usepackage[
  top=2.5cm, bottom=2.8cm,
  left=2.5cm, right=2.5cm
]{geometry}

\usepackage{amsmath}
\usepackage{amssymb}

\usepackage{graphicx}
\usepackage{float}
\usepackage{subcaption}
\usepackage[
  labelfont=bf, font=small,
  justification=justified, singlelinecheck=false
]{caption}

\usepackage{booktabs}
\usepackage{multirow}
\usepackage{array}
\usepackage[table,dvipsnames]{xcolor}
\usepackage{fontawesome5}
\usepackage{tabularx}
\usepackage{longtable}
\usepackage{ltcaption}
\usepackage{pdflscape}
\newcolumntype{C}{>{\centering\arraybackslash}X}
\newcolumntype{L}{>{\raggedright\arraybackslash}X}

\definecolor{brandcolor}{HTML}{0057B8}
\definecolor{lightlinecolor}{HTML}{CCCCCC}
\definecolor{accentgray}{HTML}{666666}
\definecolor{footergray}{HTML}{888888}
\definecolor{rowgray}{HTML}{F5F5F5}

\usepackage{listings}
\usepackage{enumitem}
\setlist[itemize]{leftmargin=1.5em, topsep=3pt, itemsep=2pt, parsep=0pt}
\setlist[enumerate]{leftmargin=1.5em, topsep=3pt, itemsep=2pt, parsep=0pt}
\usepackage{url}

\usepackage[backend=biber, style=authoryear, sorting=nyt]{biblatex}
\usepackage[
  colorlinks=true,
  linkcolor=brandcolor,
  citecolor=brandcolor,
  urlcolor=brandcolor
]{hyperref}

\usepackage{fancyhdr}
\usepackage{titlesec}

\titleformat{\section}
  {\normalfont\large\bfseries}{\thesection.}{0.5em}{}
\titlespacing{\section}{0pt}{18pt plus 3pt minus 2pt}{6pt}

\titleformat{\subsection}
  {\normalfont\normalsize\bfseries}{\thesubsection.}{0.5em}{}
\titlespacing{\subsection}{0pt}{12pt plus 2pt minus 1pt}{4pt}

\titleformat{\subsubsection}[runin]
  {\normalfont\normalsize\bfseries}{}{0em}{}[.\quad]
\titlespacing{\subsubsection}{0pt}{8pt plus 1pt}{0pt}

\fancypagestyle{firstpage}{
  \fancyhf{}
  
  \fancyfoot[L]{\footnotesize\textcolor{footergray}{\textcopyright~\the\year~Perle.ai. All rights reserved.}}
  \fancyfoot[R]{\small\textcolor{footergray}{1}}
}

\renewenvironment{abstract}{%
  \small\bfseries\noindent\ignorespaces
}{\par\medskip}

\newcommand{\brandrule}{%
  \noindent\textcolor{brandcolor}{\rule{\linewidth}{1.5pt}}\par%
}
\newcommand{\lightrule}{%
  \noindent\textcolor{lightlinecolor}{\rule{\linewidth}{0.4pt}}\par%
}

\newcommand{\apicode}[1]{\texttt{#1}}

\newcommand{\rtltext}[1]{{\arabicfont\RL{#1}}}
\newcommand{\rtlfarsi}[1]{{\arabicfont\RL{#1}}}

\newcommand{\rtlcell}[1]{{\raggedleft\arabicfont\small\setlength{\emergencystretch}{1em}\beginR #1\endR\par}}
\def\textdir TRT#1{\RL{#1}}

\usepackage{bidi}
\begin{document}
\thispagestyle{firstpage}

% ----------------------------------------------------------
% TITLE BLOCK
% ----------------------------------------------------------
\begin{flushleft}
  \includegraphics[height=20pt]{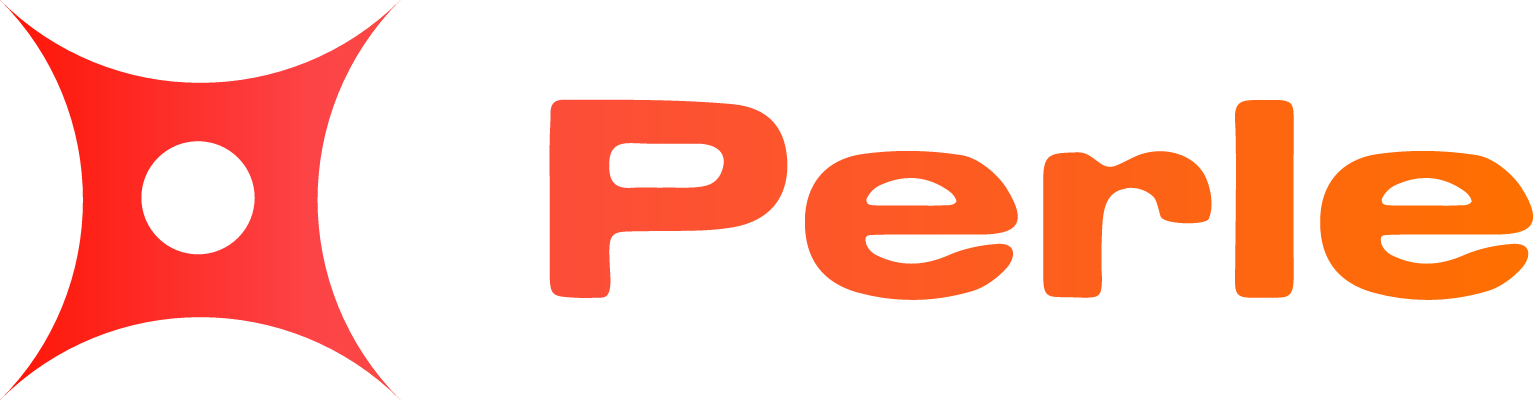}%
  \hfill
  \textcolor{accentgray}{\small May 2026}
\end{flushleft}

\vspace{4pt}
\brandrule
\vspace{8pt}

\noindent{\LARGE\bfseries Benchmarking Commercial ASR Systems on\\[2pt]
Code-Switching Speech: Arabic, Persian, and German\par}

\vspace{10pt}

\noindent
\textbf{Sajjad Abdoli}\textsuperscript{*}\footnote{Corresponding author: \href{mailto:sajjad@perle.ai}{\texttt{sajjad@perle.ai}}},\quad
\textbf{Ghassan Al-Sumaidaee}\textsuperscript{*}\footnote{ORCID: \href{https://orcid.org/0000-0002-5536-0252}{0000-0002-5536-0252}},\quad
\textbf{Clayton W. Taylor}\textsuperscript{*}\footnote{ORCID: \href{https://orcid.org/0009-0006-6478-8994}{0009-0006-6478-8994}},\quad
\textbf{Ahmad ElShiekh}\textsuperscript{*}\footnote{ORCID: \href{https://orcid.org/0009-0001-6837-6202}{0009-0001-6837-6202}},\quad
\textbf{Ahmed Rashad}\textsuperscript{*}
\par\vspace{3pt}
{\small\textsuperscript{*}Perle AI\quad
\texttt{sajjad@perle.ai}\quad
\texttt{ghassan.al-sumaidaee@perle.ai}\quad
\texttt{clayton@perle.ai}\\[2pt]
\texttt{mad.elshiekh@perle.ai}\quad
\texttt{ahmed@perle.ai}}

\vspace{10pt}

% =================================================================
\begin{abstract}
% =================================================================
Code-switching --- the natural alternation between two languages
within a single utterance --- remains one of the most challenging
and under-studied conditions for automatic speech recognition (ASR).
We present a benchmark evaluating five commercial ASR providers
across four language pairs: Egyptian Arabic--English, Saudi Arabic
(Najdi/Hijazi)--English, Persian (Farsi)--English, and German--English,
comprising 300 samples per pair selected by a two-stage pipeline
combining heuristic filtering with a GPT-4o and Gemini 1.5 Pro
ensemble scorer, reducing LLM costs by $\approx$91\%.
We evaluate on both WER and BERTScore, showing that while both
metrics agree on the ordinal ranking of systems for all Arabic and
Persian pairs ($\tau = 1.0$), WER inflates the magnitude of quality
gaps by approximately 3$\times$ by penalising semantically correct
transliteration choices. ElevenLabs Scribe v2 achieves the lowest
WER (13.2\% overall) and leads on BERTScore (0.936 overall).
Difficulty-stratified analysis reveals performance gaps masked by
aggregate averages, and BERT embedding projections confirm semantic
proximity between reference and hypothesis despite surface-level
script differences. The dataset is publicly available at
\url{https://huggingface.co/datasets/Perle-ai/ASR_Code_Switch}.
\end{abstract}

\vspace{6pt}
\lightrule
\vspace{10pt}

% =================================================================
\section{Introduction}
\label{sec:intro}
% =================================================================

Multilingual speakers rarely confine themselves to a single language.
An Egyptian software engineer on a video call might say:

\begin{quote}
\rtltext{أنا بحب الـ \LR{style} بتاعي يكون \LR{simple}}\\
\textit{(``I like my style to be simple'')}
\end{quote}

A Persian developer describing a technical problem might say:

\begin{quote}
\rtlfarsi{این \LR{feature} جدید کلی \LR{bug} داره}\\
\textit{(``This new feature has a lot of bugs'')}
\end{quote}

A Saudi product manager addressing their team:

\begin{quote}
\rtltext{الـ \LR{deadline} بكره، لازم نـ\LR{ship} الـ \LR{update}}\\
\textit{(``The deadline is tomorrow, we need to ship the update'')}
\end{quote}

These utterances are representative of everyday speech in
multilingual workplaces across the Middle East and beyond.
This phenomenon is known as \emph{code-switching} (CS):
the fluid, unconscious alternation between two languages within
a single utterance, sentence, or even morphological unit
\cite{myers1997duelling,auer2013code}.

Despite its prevalence, code-switching remains largely absent from
commercial ASR evaluation. Providers typically report WER on clean,
monolingual held-out sets. This tells practitioners almost nothing
about how a system will behave when deployed for genuinely
multilingual populations --- which today includes a substantial
portion of global knowledge workers in the Middle East, South Asia,
and Europe.

This paper makes the following contributions:

\begin{enumerate}[leftmargin=*, label=(\roman*)]
  \item A curated benchmark of 1{,}200 code-switching utterances
        (300 per language pair) across four linguistically diverse
        language pairs, selected by a reproducible two-stage pipeline.
  \item A comparative evaluation of five commercial ASR systems
        on this benchmark, reported on both WER and BERTScore.
  \item An argument and empirical demonstration that BERTScore
        is a more reliable primary metric than WER for Arabic and
        Persian code-switching due to transliteration variance.
        While WER and BERTScore agree on the ordinal ranking of
        systems ($\tau = 1.0$ for all Arabic and Persian pairs),
        WER systematically overstates the magnitude of performance
        differences by penalising semantically correct
        transliteration choices as errors.
  \item A BERT embedding space analysis providing direct visual
        evidence that semantically equivalent utterances are
        proximate regardless of script convention.
\end{enumerate}

% =================================================================
\section{Related Work}
\label{sec:related}
% =================================================================

Code-switching has been studied extensively in linguistics
\cite{gumperz1982discourse,poplack1980sometimes}, but work on
ASR for code-switching lags considerably behind monolingual
recognition. Early work focused on language identification
preceding recognition \cite{shia2004language}, while more recent
approaches integrate end-to-end models with multilingual training
data \cite{watanabe2017hybrid,toshniwal2018multilingual}.

Benchmark datasets for code-switching ASR include SEAME
(Singaporean Mandarin--English,
\cite{lyu2010seame}, the Miami
Bangor corpus \cite{deuchar2014bangor}, and the MUCS shared task
corpora \cite{diwan2021multilingual}. More recently,
\textcite{xie2026switchlingua} introduced SwitchLingua, a large-scale
multilingual and multi-ethnic code-switching benchmark with synthesized
text and recorded speech across 12 languages. Our work is
complementary but differs in focus: rather than constructing a broad
multilingual dataset, we evaluate commercial ASR systems on language
pairs and regional varieties that remain underrepresented in existing
benchmarks, including Persian--English and multiple Arabic--English
settings spanning Egyptian and Saudi dialectal speech. Arabic--English
code-switching corpora are rarer; notable exceptions include work by
\textcite{hamed2017collection} and the ArzEn corpus
\cite{hamed-etal-2020-arzen}. Persian--English and Gulf Arabic--English
code-switching datasets for ASR evaluation remain scarce.

Evaluation is another recurring challenge in code-switching ASR. WER is the dominant ASR evaluation metric \cite{makhoul1999performance},
computing the minimum edit distance between reference and hypothesis
word sequences \cite{levenshtein1966binary}. Its limitations for
code-switching include insensitivity to script-equivalent translations
and no partial credit for semantically correct transcriptions in a
different script convention. BERTScore \cite{zhang2019bertscore} addresses some of these
limitations by computing token-level cosine similarity in a
pretrained embedding space. For multilingual evaluation,
\texttt{bert-base-multilingual-cased} \cite{devlin2019bert} maps
text from 104 languages into a shared latent space, enabling
meaningful comparison across scripts.

Commercial ASR evaluation is also underdeveloped for code-switching. Several studies have benchmarked commercial ASR providers on
specific domains or noise conditions
\cite{radford2023robust,baevski2020wav2vec}, but systematic
comparison on code-switching audio across multiple language pairs
is, to our knowledge, absent from the literature.

% =================================================================
\section{Benchmark Construction}
\label{sec:benchmark}
% =================================================================
\subsection{Source Datasets}

Our source datasets are conversational transcripts collected from
real-world multilingual speakers across four language pairs: Saudi
Arabic--English, Egyptian Arabic--English, Persian (Farsi)--English,
and German--English. The candidate-selection counts are reported in
Table~\ref{tab:stage1}. Each dataset is reduced to a final benchmark
split of 300 challenging utterances. Each sample consists of a single utterance,
a reference transcript, and a corresponding audio filename. The final
300 samples for each language pair are selected from the Stage~2
scored pool by ranking rows according to \texttt{LLM Ensemble\_Score},
with \texttt{H\_Score} used as a tie-breaker. The sample-selection
protocol is detailed below.

\subsection{Voice Actor Recruitment and Recording}
\label{sec:voice_actors}

The audio recordings paired with each transcript were produced by
native speakers recruited through our internal contributor network.
For each language pair, we sourced speakers who satisfied three
criteria: (i) native or near-native fluency in the target
non-English language, with the relevant regional variety (Egyptian,
Najdi/Hijazi, Tehran-variety Farsi, or standard German);
(ii) professional or conversational fluency in English sufficient
to produce natural intra-sentential switching rather than read-aloud
pronunciation; and (iii) familiarity with the technical and
professional registers that dominate the corpus, since natural
delivery of terms such as \textit{deployment}, \textit{bottleneck},
or \textit{deadline} embedded in a matrix-language utterance
requires the speaker to use them habitually rather than to decode
them on the page.

Speakers were given the reference transcripts and instructed to
deliver each utterance conversationally rather than as a read
performance, with latitude to adjust prosody, hesitations, and minor
disfluencies to match how they would naturally produce the utterance
in context. Recordings were captured in quiet indoor environments
using consumer-grade headset or USB microphones, mirroring the
acoustic conditions of the remote-work scenarios that the benchmark
is intended to represent rather than studio-quality conditions that
would overstate ASR performance in deployment. This decision was
informed by prior corpus collection work where studio-grade audio
was observed to materially understate the error rates that the same
systems exhibited on real customer traffic; matching the recording
environment to the deployment environment is a precondition for
benchmark results that transfer. All contributors consented to the
use of their recordings for research and benchmarking purposes under
our standard contributor agreement. Per-speaker demographic details
and recruitment channel specifics are withheld to protect contributor
privacy.

\begin{figure}[ht]
\centering
\includegraphics[width=0.75\linewidth]{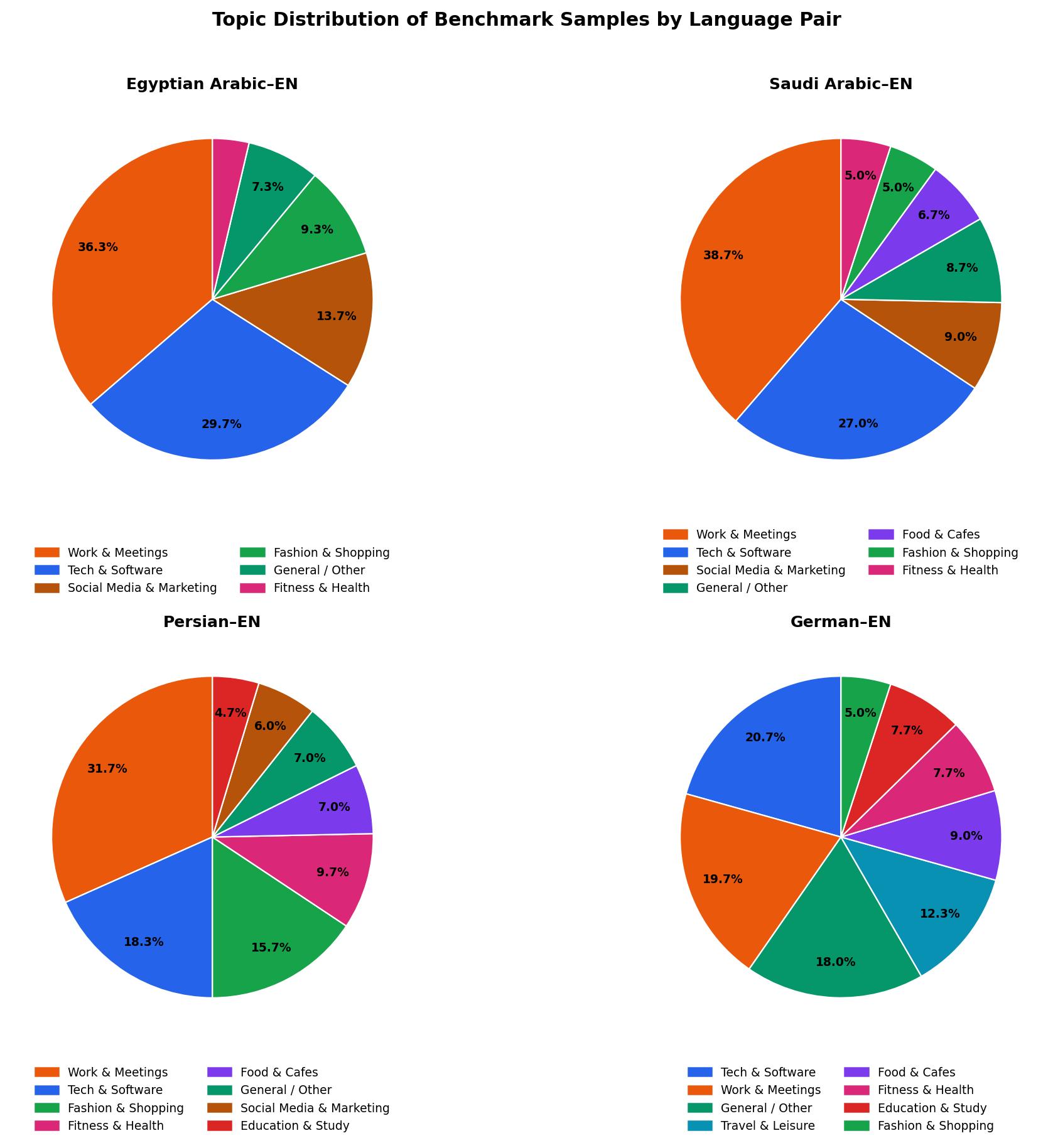}
\caption{Topic distribution of the 300 benchmark samples per language
  pair, classified by GPT-4o using an inductively derived taxonomy.
  Topics reflect the actual semantic domains of the code-switching
  speech corpora.}
\label{fig:topic_dist}
\end{figure}

Figure~\ref{fig:topic_dist} shows the distribution of semantic topics
across the 300 benchmark samples for each language pair, classified by
GPT-4o using a topic taxonomy derived inductively from the corpus itself
(no predefined categories). The taxonomy was discovered by presenting a
stratified sample of 240 transcripts to the model and asking it to
identify recurring semantic domains; all 1{,}200 transcripts were then
classified against the resulting label set.

\subsection{Stage 1: Heuristic Filter}
\label{sec:stage1}

\paragraph{Design scope.}
Stage 1 is explicitly designed for \textbf{Arabic--English and
Persian--English} pairs. Its core signals rely on the presence of
two visually distinct scripts in the same utterance: Arabic/Persian
Unicode characters (U+0600--U+06FF) and Latin characters (A--Z,
a--z). German and English share the Latin script entirely, so
script-based signals are structurally inapplicable to the German
dataset. The implications of this design choice are discussed at
the end of this subsection.

\paragraph{Scoring signals.}
Each transcript receives a composite \textbf{H\_Score} in $[0, 10]$
from five weighted signals.

\paragraph{Script mix ratio} ($w = 0.30$).
Let $n_a$ be the count of Arabic/Persian Unicode characters and
$n_l$ the count of Latin characters. The raw mix ratio is:
\begin{equation}
  m = \frac{\min(n_a,\, n_l)}{n_a + n_l}
\end{equation}
This ratio peaks at $m = 0.50$ when the two scripts appear in equal
proportion, but corpus inspection of our Arabic--English and
Persian--English datasets showed that the median script-mix ratio
in genuinely code-switched utterances falls around $0.35$: speakers
typically embed shorter English terms into longer Arabic or Persian
sentences, producing an asymmetric mix rather than a 50/50 split.
Setting $m^* = 0.35$ therefore aligns the peak of the signal with
the empirically most common mixing pattern in the data:
\begin{equation}
  h_\text{mix} = \min\!\left(\frac{m}{m^*},\; 1\right) \times 10
\end{equation}
A transcript with $m < 0.35$ scores proportionally; one with
$m \geq 0.35$ scores the maximum 10.

\paragraph{Token alternation rate} ($w = 0.30$).
Each token is classified as \textit{arabic}, \textit{latin},
or \textit{mixed} based on the majority character class.
Let $k$ be the number of consecutive token pairs that switch
between Arabic/Persian and Latin script. Then:
\begin{equation}
  h_\text{alt} = \min\!\left(\frac{k}{n/2},\; 1\right) \times 10
\end{equation}
where $n$ is the total token count. The denominator $n/2$ represents
the theoretical maximum number of switches in a sentence of $n$
tokens (every adjacent pair alternates scripts). A switch every two
tokens gives the maximum score of 10.

\paragraph{Morphological blend detection} ($w = 0.20$).
Regular expressions detect cross-language morphological fusion:
the Arabic definite article \rtltext{ال} prefixed to a Latin word
(\textit{e.g.}, \rtltext{الـ\LR{meeting}}), English stems bearing Arabic
plural or adjectival suffixes (\textit{e.g.},
developers-\rtltext{ين}), and English stems with German
derivational suffixes (\textit{e.g.}, Startup-\textit{ung}).
Let $b$ be the total regex hits:
\begin{equation}
  h_\text{morph} = \min\!\left(\frac{b}{3},\; 1\right) \times 10
\end{equation}
The ceiling of $b = 3$ was chosen empirically: manual inspection of
300 randomly sampled transcripts showed that utterances with three or
more distinct morphological blending events are already at the upper
tail of complexity, and further blending events add negligible
additional discriminative value. An utterance with one hit scores
3.3, two hits scores 6.7, and three or more hits scores the
maximum 10.

\paragraph{Length penalty} ($w = 0.10$).
Transcripts with fewer than five tokens cannot exhibit meaningful
switching and receive $h_\text{len} = 0$:
\begin{equation}
  h_\text{len} =
  \begin{cases}
    0 & \text{if } n < 5 \\
    \min\!\left(\dfrac{n - 5}{20},\; 1\right) \times 10 & \text{otherwise}
  \end{cases}
\end{equation}
The ceiling of 25 tokens ($n - 5 = 20$) reflects the 90th percentile
of utterance length in our datasets; transcripts at or above this
length receive the full score of 10.

\paragraph{Vocabulary diversity} ($w = 0.10$).
Let $|\text{vocab}|$ denote the number of \emph{unique} tokens in
the transcript (the type count), and $n$ the total token count
(the token count). Their ratio $|\text{vocab}|/n$ is the
type--token ratio (TTR), a standard measure of lexical richness.
A TTR of 1.0 means every token is unique (maximally diverse);
a TTR near 0 indicates heavy repetition. The threshold $0.7$ was
selected as the 75th-percentile TTR in our source corpora: most
naturally diverse code-switched utterances exceed this value, while
repetitive or formulaic utterances fall below it:
\begin{equation}
  h_\text{vocab} = \min\!\left(\frac{|\text{vocab}| / n}{0.7},\; 1\right)
  \times 10
\end{equation}

The composite H\_Score is:
\begin{equation}
  \label{eq:hscore}
  H = 0.30\,h_\text{mix}
    + 0.30\,h_\text{alt}
    + 0.20\,h_\text{morph}
    + 0.10\,h_\text{len}
    + 0.10\,h_\text{vocab}
\end{equation}

Table~\ref{tab:hscore_examples} shows H\_Score computation for two
representative samples from the Egyptian Arabic--English dataset,
illustrating how the five signals combine. Sample~A (H\_Score~8.37) exhibits dense mixing with six script
alternations, multiple Arabic-prefixed English tokens, and a
near-maximal mix ratio, placing it in the top difficulty quartile.
Sample~B (H\_Score~5.54) contains only one morphological blend and
fewer switches, placing it in the second quartile.

\begin{table}[ht]
\centering
\caption{H\_Score of examples (Egyptian Arabic--English).
  Sample~A is a densely code-switched tech utterance;
  Sample~B is a lighter, shorter mix.}
\label{tab:hscore_examples}
\setlength{\tabcolsep}{5pt}
\begin{tabularx}{\linewidth}{lXX}
\toprule
& \textbf{Sample A} & \textbf{Sample B} \\
\midrule
Transcript
  & \rtltext{أنا شايف إن الـ \LR{UX flow} ده محتاج \LR{redesign}، الـ \LR{user} بيـ\LR{confuse} بين الـ \LR{screens}}
  & \rtltext{اتصل بيا على الـ \LR{mobile} بكره} \\
Translation
  & \textit{``I think this UX flow needs a redesign, the user gets confused between the screens''}
  & \textit{``Call me on the mobile tomorrow''} \\
\midrule
$n$ (tokens)         & 14   & 7 \\
$n_a$ (Ar.\ chars)   & 42   & 19 \\
$n_l$ (Lat.\ chars)  & 31   & 6 \\
$m = \min/\text{sum}$& $31/73 = 0.42$ & $6/25 = 0.24$ \\
$h_\text{mix}$       & $\min(0.42/0.35,1)\times10 = 10.0$ & $\min(0.24/0.35,1)\times10 = 6.9$ \\
$k$ (switches)       & 6    & 2 \\
$h_\text{alt}$       & $\min(6/7,1)\times10 = 8.6$ & $\min(2/3.5,1)\times10 = 5.7$ \\
$b$ (morph hits)     & 2 (\rtltext{الـ\LR{UX}}, \rtltext{الـ\LR{user}}) & 1 (\rtltext{الـ\LR{mobile}}) \\
$h_\text{morph}$     & $\min(2/3,1)\times10 = 6.7$ & $\min(1/3,1)\times10 = 3.3$ \\
$h_\text{len}$       & $\min(9/20,1)\times10 = 4.5$ & $\min(2/20,1)\times10 = 1.0$ \\
$|\text{vocab}|/n$   & $14/14=1.0$ & $7/7=1.0$ \\
$h_\text{vocab}$     & $\min(1.0/0.7,1)\times10=10.0$ & $\min(1.0/0.7,1)\times10=10.0$ \\
\midrule
\textbf{H\_Score}
  & $0.30(10.0)+0.30(8.6)+0.20(6.7)+0.10(4.5)+0.10(10.0) = \mathbf{8.37}$
  & $0.30(6.9)+0.30(5.7)+0.20(3.3)+0.10(1.0)+0.10(10.0) = \mathbf{5.54}$ \\
\bottomrule
\end{tabularx}
\end{table}

\paragraph{Per-dataset candidate selection.}
Table~\ref{tab:stage1} shows how many rows are passed from Stage 1
to Stage 2 per dataset. For Saudi, a random pre-sample of 5{,}000
rows is drawn from the 27{,}190-row source corpus before heuristic
scoring to manage computational cost.

\begin{table}[ht]
\centering
\caption{Stage 1 candidate selection parameters.}
\label{tab:stage1}
\begin{tabular}{lrrrl}
\toprule
\textbf{Dataset} & \textbf{Source} & \textbf{Pre-sample} &
  \textbf{LLM Candidates} & \textbf{Filter active?} \\
\midrule
Saudi    & 27{,}190 & 5{,}000 & 1{,}500 & Yes ($\sim$70\% reduction) \\
Egyptian &  9{,}153 & none    & 1{,}200 & Yes ($\sim$87\% reduction) \\
Persian  &  1{,}934 & none    &   600   & Yes ($\sim$69\% reduction) \\
German   &    860   & none    &   860   & No --- all rows forwarded \\
\bottomrule
\end{tabular}
\end{table}

\paragraph{Stage 1 and the German--English pair.}
The heuristic filter is designed for Arabic--English and
Persian--English, where code-switching is structurally visible as
an alternation between two distinct Unicode script blocks.
For German--English, both languages share the Latin script, so
$h_\text{mix} = 0$ and $h_\text{alt} = 0$ identically for all
samples. The effective weight sum is therefore $W_\text{eff} = 0.40$
and $H \in [0, 4]$ rather than $[0, 10]$. This compressed range has
no practical consequence since all 860 German rows are forwarded to
Stage 2 unconditionally, and H\_Score serves only as a tie-breaker
rather than a selection threshold. Consumers of H\_Score as a
standalone difficulty indicator should note that German scores are
not directly comparable to Arabic or Persian scores on the
$[0, 10]$ scale.

The substantive work for German--English is done entirely by the
LLM ensemble in Stage 2. GPT-4o and Gemini 1.5 Pro assess
German--English code-switching on its own linguistic merits ---
intra-sentential lexical switching, German derivational morphology
applied to English stems (\textit{e.g.}, \textit{downloaden},
\textit{gedownloadet}), register mixing between colloquial German
and English technical vocabulary, and phonological ambiguity at
the German--English boundary --- without any dependence on script
detection. The LLM ensemble is therefore the appropriate and
sufficient tool for this language pair, and the quality of the
resulting German benchmark reflects Stage 2 scoring alone.

\subsection{Stage 2: LLM Ensemble Scoring}
\label{sec:stage2}

Stage 1 captures surface-level structural difficulty for Arabic and
Persian pairs and forwards all German rows unconditionally.
Stage 2 captures semantic and phonological complexity that requires
genuine linguistic reasoning for all four language pairs.

Each Stage 1 candidate is sent concurrently to \textbf{GPT-4o}
and \textbf{Gemini 1.5 Pro} (one thread per model). Each model
scores the transcript across six linguistic dimensions on a 1--10
scale (Table~\ref{tab:dimensions}), returns verbatim evidence per
dimension, up to five annotated hard tokens with one-sentence ASR
risk explanations, and a summary sentence.

\begin{table}[ht]
\centering
\caption{LLM scoring dimensions (Stage 2).}
\label{tab:dimensions}
\begin{tabularx}{\linewidth}{lX}
\toprule
\textbf{Dimension} & \textbf{What it captures} \\
\midrule
Morphological blending      & Cross-language affixes on foreign-language stems \\
Switching density           & Intra-sentential language switch frequency \\
Slang and register mix      & Colloquial dialect forms meeting domain-specific jargon \\
Phonological ambiguity      & Tokens acoustically plausible in either language \\
Named entity/jargon density & Brands, technical terms, proper nouns in mixed context \\
Script/orthographic complexity & Romanisation inconsistency, mixed-script tokens \\
\bottomrule
\end{tabularx}
\end{table}

The \textbf{Ensemble\_Score} is the average of both models'
\texttt{overall\_score} values (1--10). Using two independent
models reduces the influence of any single model's idiosyncratic
biases; rows where the two models disagree by more than 3 points
on any dimension are flagged for inspection.

The final 300 samples per language pair are ranked by
Ensemble\_Score (descending), with H\_Score as a tie-breaker.
The pipeline is fully resumable via SQLite checkpoint. This
two-stage design reduces LLM API calls by approximately 91\%
compared to scoring all rows directly for Arabic and Persian
pairs; for German, all rows are LLM-scored since the dataset
size makes exhaustive scoring practical at negligible cost.

% =================================================================
\section{Evaluation Methodology}
\label{sec:eval}
% =================================================================

\subsection{Systems Evaluated}

We evaluate five commercial ASR systems accessed via their production
APIs in May 2026. Table~\ref{tab:systems} summarises the models;
Table~\ref{tab:api_params} lists the exact API parameters used in
our benchmark.

\noindent\textbf{Key architectural distinctions.}
The five systems represent three different strategies for handling
code-switching:

\begin{itemize}
\item \textbf{Language-agnostic end-to-end models}
        (ElevenLabs Scribe v2, Google Chirp 3):
        ElevenLabs Scribe v2 is a state-of-the-art speech
        recognition model supporting 90+ languages with
        \textit{smart language detection} --- no language
        code is required; the model infers the spoken language
        directly from the audio~\cite{elevenlabs2025scribev2}.
        Google Chirp 3 is the latest generation of Google's
        multilingual ASR-specific generative models, providing
        enhanced transcription accuracy with automatic language
        detection; setting \apicode{language\_codes=["auto"]}
        instructs the model to ``automatically infer and transcribe
        in the most prevalent language'' without any candidate
        list~\cite{google2025chirp3}.
        Neither model is given a language hint in our benchmark,
        making this the most demanding configuration for
        code-switching: the model must resolve script and language
        identity from acoustic evidence alone, without the
        disambiguation advantage of a pre-specified locale set.

\item \textbf{LLM-integrated decoder} (OpenAI gpt-4o-transcribe):
        the model builds upon the GPT-4o architecture and is
        extensively pre-trained on specialised audio-centric
        datasets~\cite{openai2025transcribe}. Rather than the
        conventional Whisper-style encoder-decoder trained
        primarily on supervised speech data, gpt-4o-transcribe
        adopts what OpenAI describes as a ``reinforcement
        learning (RL)-heavy paradigm'' for transcription,
        combined with extensive midtraining on diverse,
        high-quality audio~\cite{openai2025transcribe}.
        This training regime enables the model to leverage the
        GPT-4o backbone's language priors to resolve ambiguity
        at script and language boundaries --- a property
        particularly relevant for code-switching audio where
        acoustic evidence alone is insufficient to determine
        the correct script convention. No language hint is
        supplied in our benchmark configuration.

\item \textbf{Segment-level language identification}
        (Azure Speech CLID): a discrete language identification
        (LID) classifier runs at each recognition segment boundary,
        re-estimating the active language before the acoustic
        decoder produces the transcript for that segment.
        Per Microsoft's documentation, this architecture
        \textit{``doesn't support changing languages within the
        same sentence''}: intra-sentential switches are invisible
        to the LID layer and are decoded under whichever language
        was detected at the segment onset~\cite{microsoft2024azurelid}.
        We set
        \apicode{SpeechServiceConnection\_LanguageIdMode=Continuous}
        explicitly via the Python SDK, which triggers re-identification
        at each segment boundary rather than once per session
        (the \apicode{AtStart} default); this alone improved mean
        WER by 2.0 percentage points across all four language pairs.

\item \textbf{Unified multilingual decoder with implicit
        language handling} (Deepgram Nova-3): unlike Azure's
        two-stage design, Nova-3 performs no explicit LID step.
        Per Deepgram's technical documentation, the model
        ``operates as a truly unified multilingual speech
        recognition system'' that ``naturally emits transcriptions
        following the speaker's language switches, without relying
        on explicit routing or language-specific
        mechanisms''~\cite{deepgram2025nova3}. This is achieved
        through ``a multi-stage training process that combines
        synthetic code-switched data at massive scale with
        carefully curated real-world datasets''~\cite{deepgram2025nova3},
        covering 10 languages: English, Spanish, French, German,
        Hindi, Russian, Portuguese, Japanese, Italian, and
        Dutch~\cite{deepgram2025nova3multi}.
        Arabic and Persian are absent from this language list,
        which is why Nova-3 results are suppressed for all
        Arabic and Persian pairs in our benchmark.
\end{itemize}

This architectural distinction is consequential for code-switching
evaluation: language-agnostic and LLM-integrated decoders can in
principle produce mixed-script output within a single sentence,
whereas segment-level LID systems are structurally limited to
producing one script per recognition segment. The WER and BERTScore
results reported in Section~\ref{sec:results} reflect this
constraint directly.

Deepgram Nova-3 does not list Arabic or Persian in its documented
CS language support. For these pairs we suppress WER and BERTScore
and exclude the system from aggregate rankings --- evaluating a
system outside its design envelope produces numbers that reflect
the absence of capability, not a fair comparison.

\subsection{Text Normalisation}

Before computing any metric, both reference and hypothesis are
normalised: lowercased, punctuation removed, whitespace collapsed.
\textbf{Arabic and Persian script characters are preserved
throughout} --- a model transcribing Arabic-script speech in Latin
characters is penalised, not rewarded.

We note that production Arabic and Persian ASR pipelines typically
apply additional canonicalisation steps beyond what we use here:
unification of hamza variants
(\rtltext{أ}/\rtltext{إ}/\rtltext{ا}),
collapsing \textit{ta marbuta} to \textit{ta}
(\rtltext{ة}~$\to$~\rtltext{ه}),
diacritic stripping, and conversion between Arabic-Indic
(\rtltext{٠١٢}) and Western (012) digit forms. We deliberately omit
these here to evaluate the raw output of each provider's decoder
without masking systematic preprocessing differences between
systems. A consequence is that orthographic variants that a
production system would treat as identical are penalised by WER in
our setup. BERTScore is partially robust to these variants because
the multilingual encoder maps them to nearby points in embedding
space; this is an additional reason to report both metrics rather
than either alone.

\subsection{Word Error Rate}

WER is defined as:
\begin{equation}
  \label{eq:wer}
  \text{WER} = \frac{S + D + I}{N}
\end{equation}
where $S$, $D$, $I$ are substituted, deleted, and inserted words in
the minimum-edit-distance alignment between reference (length $N$)
and hypothesis, found via dynamic programming (the \texttt{jiwer}
library). WER can exceed 1.0 when insertions are numerous.

Crucially, WER has no concept of semantic equivalence: the
substitution \texttt{feature}~$\to$~\rtlfarsi{فیچر} (Persian
transliteration of ``feature'') incurs the same unit penalty as
substituting a completely wrong word.

\subsection{BERTScore}
\label{sec:bertscore}

BERTScore \cite{zhang2019bertscore} computes soft token alignment
in the embedding space of a pretrained transformer.
Let $R = \{r_1, \ldots, r_m\}$ and $H = \{h_1, \ldots, h_n\}$ be
the reference and hypothesis token sets with contextual embeddings
$\mathbf{r}_i, \mathbf{h}_j \in \mathbb{R}^d$. Precision and recall:

\begin{align}
  P_\text{BERT} &= \frac{1}{|H|} \sum_{h_j \in H}
    \max_{r_i \in R}\; \cos(\mathbf{h}_j,\, \mathbf{r}_i)
  \label{eq:bert_p} \\[4pt]
  R_\text{BERT} &= \frac{1}{|R|} \sum_{r_i \in R}
    \max_{h_j \in H}\; \cos(\mathbf{r}_i,\, \mathbf{h}_j)
  \label{eq:bert_r}
\end{align}

F1 is the harmonic mean:
\begin{equation}
  F_\text{BERT} = 2 \cdot
    \frac{P_\text{BERT} \cdot R_\text{BERT}}
         {P_\text{BERT} + R_\text{BERT}}
  \label{eq:bert_f1}
\end{equation}

We use \texttt{bert-base-multilingual-cased} (mBERT;
\citeauthor{devlin2019bert} (\citeyear{devlin2019bert})),
pre-trained on 104 languages. The critical property for CS
evaluation is that mBERT maps tokens from all languages into a
\emph{shared} latent space, so semantically equivalent tokens in
different scripts --- \textit{e.g.}, \texttt{feature} and
\rtlfarsi{فیچر} --- are geometrically proximate. A transcription
semantically correct but using a different script convention
receives a high $F_\text{BERT}$, whereas it receives a full
substitution penalty under WER. We use \texttt{lang="others"}
(multilingual mode, no baseline rescaling) and batch size 32.

% =================================================================
\section{Results}
\label{sec:results}
% =================================================================

\subsection{Overall Rankings}

\begin{figure}[ht]
\centering
\includegraphics[width=0.55\linewidth]{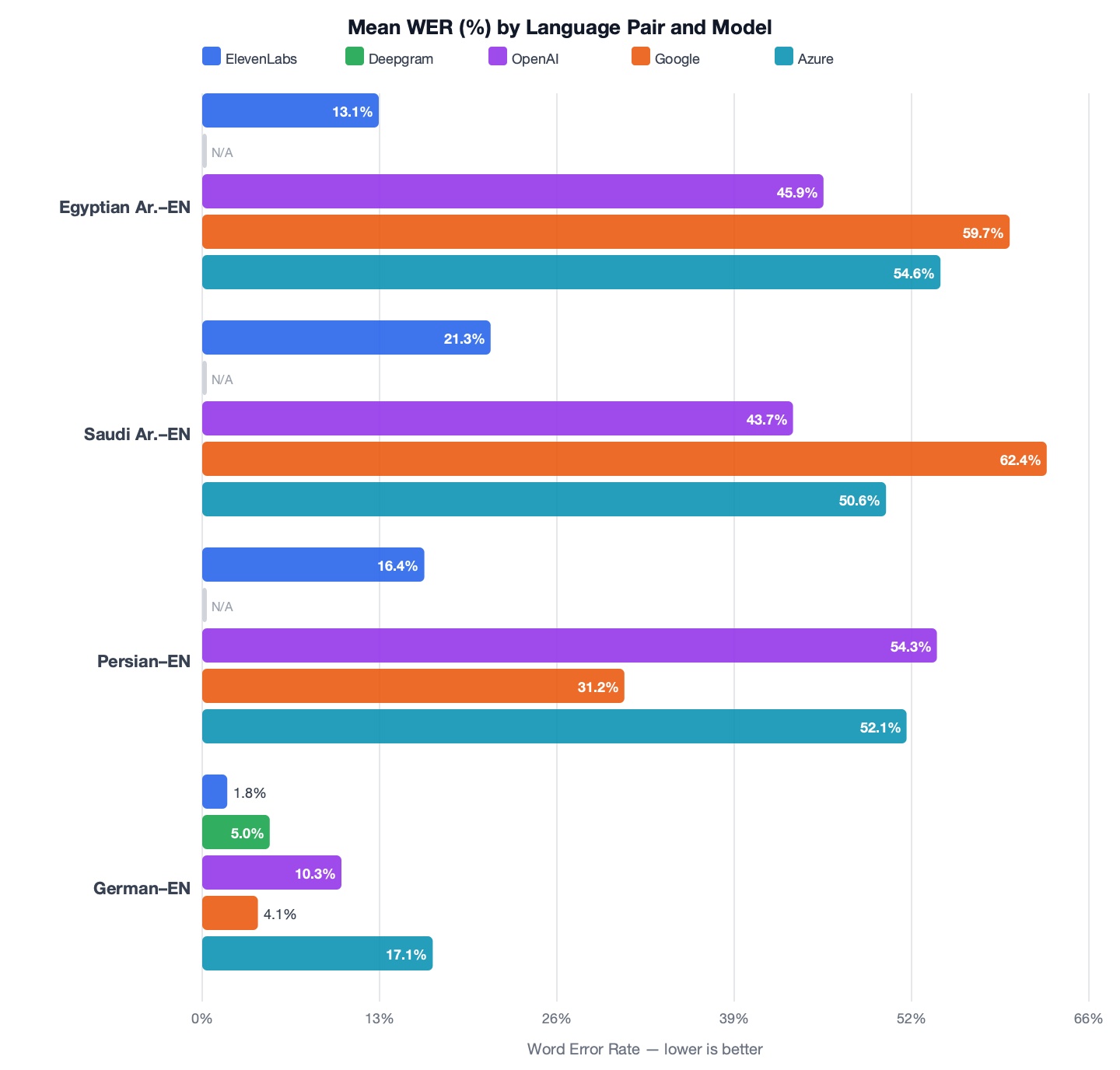}
\caption{Mean WER (\%) for all five systems across all four language
  pairs. Each group of bars represents one language pair;
  bars within a group are colour-coded by system.
  Deepgram results shown for German only (no Arabic/Persian CS support).
  Lower is better.}
\label{fig:combined_wer}
\end{figure}

\begin{figure}[ht]
\centering
\includegraphics[width=0.55\linewidth]{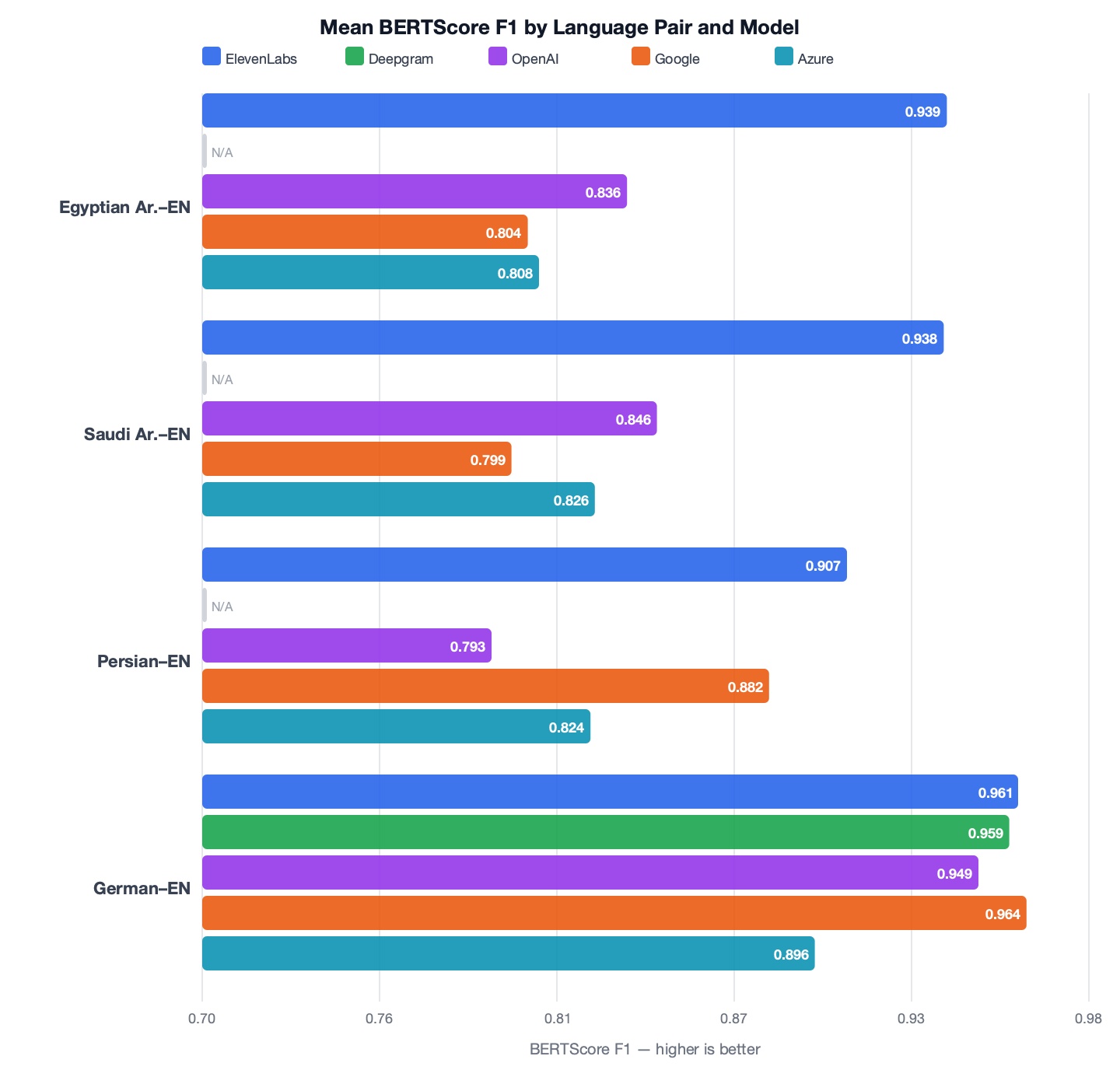}
\caption{Mean BERTScore F1 for all five systems across all four
  language pairs (\texttt{bert-base-multilingual-cased}).
  BERTScore rewards semantically correct transliteration that
  WER would penalise. Higher is better.}
\label{fig:combined_bert}
\end{figure}

Table~\ref{tab:overall} reports the overall mean WER and BERTScore
for each system. Figures~\ref{fig:combined_wer} and
\ref{fig:combined_bert} show the full breakdown across all four
language pairs in a single view. ElevenLabs Scribe v2 leads on
both metrics across all four language pairs.

\begin{table}[ht]
\centering
\caption{Overall results across CS-supported language pairs.
  Deepgram evaluated on German only.}
\label{tab:overall}
\begin{tabular}{lrrr}
\toprule
\textbf{System} & \textbf{Mean WER} & \textbf{Mean BERTScore} &
  \textbf{CS Pairs} \\
\midrule
ElevenLabs Scribe v2     & \textbf{13.2\%} & \textbf{0.936} & 4 \\
OpenAI gpt-4o-transcribe & 38.6\%          & 0.856          & 4 \\
Google Chirp 3           & 39.4\%          & 0.862          & 4 \\
Azure AI Speech (CLID)   & 43.6\%          & 0.839          & 4 \\
Deepgram Nova-3          & 5.0\%$^\dagger$ & 0.959$^\dagger$& 1 \\
\bottomrule
\multicolumn{4}{l}{$^\dagger$ German only; not directly comparable.}
\end{tabular}
\end{table}

% ── Combined WER figure ──────────────────────────────────────────

% ── Combined BERTScore figure ─────────────────────────────────────

\noindent\textbf{Note on Deepgram.}
Deepgram Nova-3 achieves 5.0\% WER and 0.959 BERTScore, but these
figures are from the German--English pair only. Its ``overall''
number reflects one of the easiest language pairs in our benchmark
and is not comparable to systems evaluated across all four pairs.

\subsection{Performance Across Language Pairs}

Figures~\ref{fig:combined_wer} and~\ref{fig:combined_bert} report the
per-language-pair results, making separate result tables unnecessary.
Across Egyptian Arabic--English, Saudi Arabic--English,
Persian--English, and German--English, ElevenLabs Scribe v2 is the
most consistent system: it achieves the lowest WER for all four
language pairs and remains competitive or leading on BERTScore.

The Arabic pairs are substantially harder than German--English.
Egyptian Arabic--English includes dense intra-sentential switching,
as in:

\begin{quote}
\rtltext{أنا شايف إن الـ \LR{UX flow} ده محتاج \LR{redesign}،
الـ \LR{user} بيـ\LR{confuse} بين الـ \LR{screens}}\\[2pt]
\textit{(``I think this UX flow needs a redesign,
the user gets confused between the screens'')}
\end{quote}

Saudi Arabic--English introduces additional dialectal variation
through Najdi and Hijazi speech. Google Chirp 3 performs worse on
Saudi Arabic than on Egyptian Arabic, suggesting that Gulf dialectal
phonology remains less robustly covered than more widely represented
Arabic varieties. ElevenLabs retains a clear advantage on both Arabic
pairs, indicating that its gains are not limited to a single dialect.

Persian--English shows the strongest mismatch between WER and
semantic adequacy. Farsi uses right-to-left Perso-Arabic script,
while English words may remain in Latin script or be transliterated.
For example, a reference containing \LR{feature}, \LR{bug}, and
\LR{deadline} may be transcribed with semantically equivalent Persian
forms such as \rtlfarsi{فیچر} and \rtlfarsi{باگ}. WER treats these as
string substitutions, whereas BERTScore gives partial credit for their
semantic equivalence. This explains why Persian results in
Figure~\ref{fig:combined_bert} are less severe than the corresponding
WER gaps in Figure~\ref{fig:combined_wer}.

German--English is the easiest pair because both languages use the
Latin script, reducing transliteration ambiguity. The leading systems
cluster closely on this pair, and WER and BERTScore are much more
concordant than for Arabic or Persian. Deepgram Nova-3 is therefore
reported only for German--English and should not be interpreted as an
overall competitor to systems evaluated across all four pairs.

Taken together, the language-pair results show that aggregate scores
hide important structure: Arabic performance is shaped by dialectal
coverage, Persian performance is shaped by script and transliteration
choices, and German performance reflects a lower-ambiguity
code-switching setting.

Appendix~\ref{app:qualitative} provides the full qualitative
side-by-side transcription comparison for the highest-divergence
examples across all language pairs, including Table~\ref{tab:qualitative_full}.

% =================================================================
\section{Analysis}
\label{sec:analysis}
% =================================================================

\subsection{Difficulty-Stratified Performance}

A well-calibrated benchmark should show rising error rates as
difficulty increases. We stratify all CS-supported samples into
quartiles by H\_Score and compute mean WER per quartile
(Table~\ref{tab:wer_quartile}).

\begin{table}[ht]
\centering
\caption{Mean WER by H\_Score difficulty quartile
  (CS-supported samples; Deepgram excluded).}
\label{tab:wer_quartile}
\begin{tabular}{lrrrr}
\toprule
\textbf{Quartile} &
  \textbf{ElevenLabs} &
  \textbf{Google} &
  \textbf{OpenAI} &
  \textbf{Azure} \\
\midrule
Q1 (easiest) & 2.0\%  & 4.4\%  & 9.7\%  & 17.1\% \\
Q2           & 13.9\% & 30.8\% & 46.2\% & 46.4\% \\
Q3           & 15.0\% & 54.4\% & 48.7\% & 54.2\% \\
Q4 (hardest) & 20.0\% & 61.5\% & 45.2\% & 52.2\% \\
\bottomrule
\end{tabular}
\end{table}

Benchmark calibration holds: WER increases Q1$\to$Q4 for all
systems. The gap between ElevenLabs and Google Chirp 3 is
2.4 percentage points at Q1 but 41.5 points at Q4. Systems that
appear broadly comparable on aggregate WER diverge dramatically
when evaluated on the hardest code-switching instances.

\subsection{BERTScore Across Difficulty Quartiles}

Table~\ref{tab:bert_quartile} reveals a counterintuitive finding:
ElevenLabs' BERTScore \emph{increases} from Q3 (0.929) to
Q4 (0.938) --- the hardest quartile receives the highest
non-Q1 score of any system.

\begin{table}[ht]
\centering
\caption{Mean BERTScore F1 by H\_Score difficulty quartile.}
\label{tab:bert_quartile}
\begin{tabular}{lrrrr}
\toprule
\textbf{Quartile} &
  \textbf{ElevenLabs} &
  \textbf{Google} &
  \textbf{OpenAI} &
  \textbf{Azure} \\
\midrule
Q1 (easiest) & 0.961 & 0.962 & 0.949 & 0.896 \\
Q2           & 0.920 & 0.889 & 0.826 & 0.839 \\
Q3           & 0.929 & 0.819 & 0.826 & 0.812 \\
Q4 (hardest) & \textbf{0.938} & 0.797 & 0.837 & 0.817 \\
\bottomrule
\end{tabular}
\end{table}

Q4 is dominated by Arabic and Persian utterances with heavy
morphological blending and dense switching. We hypothesise that
ElevenLabs consistently picks a single script convention and applies
it coherently, whereas other models produce inconsistent outputs.
WER penalises any deviation from the reference's script convention;
BERTScore rewards semantic fidelity regardless of script. ElevenLabs'
Q4 BERTScore advantage therefore suggests it is capturing semantic
content correctly even when WER would count its output as wrong.

\subsection{WER--BERTScore Divergence by Language Pair}
\label{sec:divergence}

We quantify the agreement between WER and BERTScore rankings using
Kendall's $\tau$, computed across all CS-supported systems for each
language pair. For $n$ systems there are $\binom{n}{2}$ possible
system pairs. A pair $(i, j)$ is \emph{concordant} ($C$) if both
metrics agree on which system is better, and \emph{discordant} ($D$)
if they disagree. Kendall's $\tau$ is then:
\begin{equation}
  \tau = \frac{C - D}{\binom{n}{2}}
  \label{eq:tau}
\end{equation}
$\tau = 1$ indicates perfect agreement between the two metric
rankings; $\tau = 0$ indicates no correlation.

\begin{table}[ht]
\centering
\caption{WER--BERTScore rank concordance (Kendall's $\tau$)
  per language pair, computed from the benchmark results.
  $n$ = number of CS-supported systems evaluated.}
\label{tab:divergence}
\begin{tabular}{lrrr}
\toprule
\textbf{Language Pair} & $n$ & \textbf{Pairs} &
  \textbf{Kendall's $\tau$} \\
\midrule
Egyptian Arabic--English & 4 & 6  & 1.000 \\
Saudi Arabic--English    & 4 & 6  & 1.000 \\
Persian--English         & 4 & 6  & 1.000 \\
German--English          & 5 & 10 & 0.800 \\
\bottomrule
\end{tabular}
\end{table}

Table~\ref{tab:divergence} reveals that WER and BERTScore produce
\textbf{identical system rankings} for all three Arabic and Persian
pairs ($\tau = 1.0$, all six system pairs concordant): both metrics
place ElevenLabs Scribe v2 first and Google Chirp 3 last on Egyptian
and Saudi Arabic, and both place ElevenLabs first and OpenAI last on
Persian. The transliteration variance problem does not cause metric
disagreement at the \emph{ordinal ranking} level.

The divergence is instead in the \emph{absolute score magnitudes}.
WER reports a 46.6 percentage point spread between ElevenLabs
(13.1\%) and Google Chirp 3 (59.7\%) on Egyptian Arabic;
BERTScore reports a spread of only 0.135 (0.939 vs 0.804).
WER inflates the apparent quality gap because it penalises
transliteration choices as errors, even when the hypothesis is
semantically correct. A system that transcribes
\rtltext{الـ feature} as \texttt{the feature} incurs WER
substitution penalties; BERTScore maps both to the same
neighbourhood in the multilingual embedding space and scores them
as equivalent. The practical implication is clear: for Arabic and
Persian ASR evaluation, WER correctly identifies the ranking of
systems but overstates the performance differences between them.
BERTScore should be reported alongside WER to give an accurate
picture of the magnitude of quality gaps.

German--English is the only pair where the two metrics produce a
discordant ranking ($\tau = 0.800$, nine of ten system pairs
concordant, one discordant). ElevenLabs Scribe v2 achieves lower
WER than Google Chirp 3 (1.8\% vs 4.1\%) but slightly lower
BERTScore (0.961 vs 0.964). The inversion reflects a regime
difference at near-zero error rates: at this level of accuracy,
BERTScore detects subtle semantic differences that WER, operating
on near-perfect word sequences, cannot resolve. This is consistent
with the absence of transliteration ambiguity in German--English
--- both metrics are measuring genuine transcription quality, and
they converge to within 0.003 BERTScore points while disagreeing
by 2.3 WER percentage points.

\subsection{BERT Embedding Space Analysis}
\label{sec:embedding}

To provide direct visual evidence that semantically equivalent
code-switching utterances are geometrically proximate in the
multilingual embedding space --- regardless of script convention ---
we project a sample of Persian reference and ElevenLabs hypothesis
sentence embeddings into two dimensions.

\begin{figure}[ht]
\centering
\includegraphics[width=0.55\linewidth]{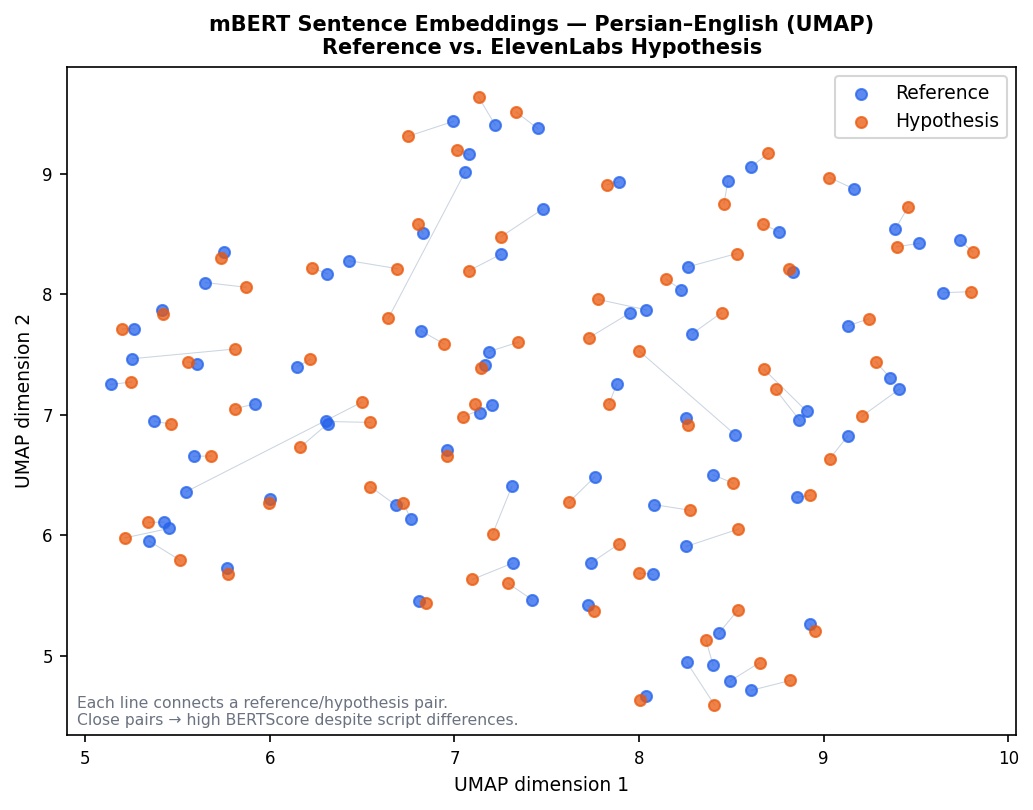}
\caption{UMAP projection of multilingual sentence embeddings for
  80 Persian--English utterance pairs. Blue = reference transcripts;
  orange = ElevenLabs Scribe v2 hypotheses. Each grey line connects
  a reference--hypothesis pair. The tight clustering of connected
  pairs --- short line lengths --- confirms semantic proximity
  despite surface-level script differences (\textit{e.g.},
  \texttt{feature} vs.\ \rtlfarsi{فیچر}), consistent with the
  high Persian--English BERTScore values reported in Section~\ref{sec:results}.
  Pairs with longer connecting lines correspond to samples where
  the model hallucinated or omitted content.}
\label{fig:embedding}
\end{figure}

We encode 80 randomly sampled Persian--English utterance pairs using
\cite{reimers2019sentencebert}, a multilingual sentence encoder
trained to place semantically equivalent sentences close together
across languages, and reduce to 2D using UMAP
\cite{mcinnes2018umap}. Figure~\ref{fig:embedding} shows the result.

Figure~\ref{fig:embedding} reveals several properties of the
multilingual embedding space that directly support our argument.
First, reference (blue) and hypothesis (orange) points are spatially
interleaved throughout the 2D projection with no visible separation
between the two clouds, confirming that the encoder does not
represent script convention as a distinguishing dimension.
Second, the majority of connecting lines are short relative to the
typical inter-point distance in the cloud, indicating that most
reference--hypothesis pairs are geometrically closer to each other
than to other utterances --- the visual correlate of high BERTScore.
Third, the loose clusters visible in the projection reflect topical
and structural similarity rather than language identity: reference
and hypothesis points from the same semantic domain co-locate
regardless of script convention.
A small number of pairs with visibly long connecting lines
($\sim$10--15\% of the sample) correspond to genuinely erroneous
transcriptions where content was hallucinated or omitted; these are
the cases where both WER and BERTScore correctly agree that the
transcription is wrong.
Crucially, there is no systematic directional offset between blue
and orange points, ruling out any encoding bias toward one script
convention. The encoder treats \texttt{feature} and
\rtlfarsi{فیچر} as the same point in meaning-space --- which is
precisely why BERTScore rewards a transcription that WER penalises.

% =================================================================
\section{Discussion}
\label{sec:discussion}
% =================================================================

\subsection{On the Choice of Evaluation Metric}

Our results provide concrete evidence that WER, while reliable for
\emph{ranking} ASR systems on Arabic and Persian code-switching,
systematically overstates the magnitude of performance differences.
The Kendall's $\tau$ analysis (Section~\ref{sec:divergence}) shows
perfect rank concordance between WER and BERTScore for all Arabic
and Persian pairs ($\tau = 1.0$): both metrics agree on which
system is best and worst. However, WER reports a 46.6 percentage
point spread on Egyptian Arabic where BERTScore reports only 0.135
--- a 3.4$\times$ inflation of the apparent quality gap. The
Persian example in Section~\ref{sec:bertscore} illustrates the
mechanism: 30.8\% WER for a semantically correct transcription
whose only ``errors'' are consistent transliteration choices.
The embedding analysis in Section~\ref{sec:embedding} confirms
this at scale.

We recommend: (1) always report both WER and BERTScore for Arabic
and Persian CS pairs; (2) treat BERTScore as the primary metric
for \emph{measuring the size of quality gaps} between systems,
since it is not inflated by transliteration variance;
(3) if script consistency is a hard requirement (\textit{e.g.},
for downstream NLP pipelines expecting a specific encoding),
supplement BERTScore with a script-normalised WER that maps all
script variants to a canonical form before comparison.

\subsection{Difficulty Stratification and Benchmark Design}

The large performance spread across difficulty quartiles
(Table~\ref{tab:wer_quartile}) highlights a general limitation
of aggregate ASR benchmarks: mean WER over a diverse corpus can
obscure failure modes that only manifest at high switching density
or morphological complexity. Our two-stage selection pipeline
deliberately over-represents these hard cases.

The rising Q1$\to$Q4 WER gradient validates the H\_Score as a
meaningful proxy for ASR difficulty despite being computed entirely
from surface-level statistics with no semantic knowledge.

\subsection{Dialect Stratification within Arabic}
\label{sec:dialect_strat}

The two Arabic--English language pairs in our benchmark
(Egyptian Arabic--English and Saudi Arabic--English) are reported
as separate aggregates, but the Saudi dataset itself spans at least
two distinct dialectal varieties --- Najdi and Hijazi --- which
differ phonetically and lexically. Aggregating across them yields
a single WER number that can mask substantial per-dialect failure
modes. This is not a hypothetical concern: in production Arabic
ASR deployments, a model that achieves acceptable mean WER across a
mixed-dialect corpus can still fail consistently on one specific
dialect, producing systematically worse experience for the segment
of customers who speak it. The aggregate number passes; the customer
experience does not.

We therefore recommend that future ASR benchmarks targeting
multi-dialect languages report dialect-stratified WER and BERTScore
as a first-class result, on the same footing as language-pair
aggregates. The same argument extends to Persian (regional varieties
beyond Tehran-variety Farsi) and to any language with substantial
internal dialectal variation. A practical consequence for benchmark
construction is that dialect coverage should be treated as a
separate sampling axis from language pair coverage: balancing
language pairs without balancing dialects within them can still
produce a benchmark that systematically underrepresents particular
speaker populations. Where per-sample dialect labels are unavailable
in the source corpus --- as is partially the case for our Saudi
dataset --- this analysis cannot be performed retroactively, and
we flag this as a corpus-design recommendation rather than a result
we are able to report here in full.

\subsection{Deployment Considerations}
\label{sec:deployment}

The benchmark results reported in this paper measure offline
transcription quality on pre-recorded audio. Practitioners selecting
an ASR provider for production deployment face a second axis of
considerations that aggregate WER and BERTScore do not capture, and
which materially affect provider selection.

\paragraph{Latency.}
LLM-integrated decoders such as gpt-4o-transcribe and large
end-to-end models such as Scribe v2 typically exhibit higher
per-request latency than streaming-optimised systems. For
batch transcription workloads (call recordings, archival audio,
research corpora) this is irrelevant. For real-time use cases
(live captioning, conversational agents, contact-centre assistance)
the highest-accuracy model on this benchmark may not be the
right choice; a system with marginally higher WER but
sub-second streaming latency can deliver a better end-user
experience.

\paragraph{Cost at production volume.}
Per-minute pricing varies by approximately an order of magnitude
across the providers evaluated. For an enterprise transcribing tens
of thousands of hours per month, this dominates the total cost of
ownership and can reverse the apparent ranking from accuracy alone.
We deliberately do not report current pricing in this paper because
provider pricing changes faster than the publication cycle, but
practitioners should compute the cost-adjusted ranking for their
specific volume profile rather than selecting on WER alone.

\paragraph{Segment-level LID failure modes in real traffic.}
The architectural limitation discussed in Section~\ref{sec:eval} ---
that segment-level LID systems cannot switch languages within a
recognition segment --- has a concrete operational consequence in
customer support traffic, where a single utterance frequently
contains a matrix-language sentence with one or two embedded
English technical terms. In our experience deploying Arabic ASR
to Gulf enterprise customers, this is the modal utterance shape in
banking, healthcare, and telco support calls, not the exception.
A system that cannot represent within-utterance switching will
systematically misrecognise this shape regardless of how well it
performs on the monolingual segments either side of the switch.

\paragraph{Evaluation harness design.}
We emphasise that the choice of evaluation methodology matters as
much as the choice of model. A benchmark that reports a single
aggregate WER number across all dialects and all switching densities
will rank models very differently from one that stratifies by
difficulty (Table~\ref{tab:wer_quartile}) and by dialect
(Section~\ref{sec:dialect_strat}). Practitioners selecting a
provider for a specific deployment should build their own
evaluation harness against samples representative of their actual
traffic, rather than relying on any single published WER figure ---
including ours.

\subsection{Future Work}

\paragraph{Broader speaker demographics.}
Future benchmark releases should expand coverage beyond the current
Saudi Najdi and Hijazi varieties to include additional Gulf dialects.
Similarly, the Persian dataset should be extended beyond informal
Tehran-variety speech to include a wider range of regional Farsi
varieties and speaker backgrounds.

\paragraph{Wider domain coverage.}
Future work should evaluate code-switching ASR outside the technology
and professional contexts emphasized here. Domains such as healthcare,
education, public services, and casual conversation may exhibit
different code-switching patterns, lower English loanword density,
and different error profiles.

\paragraph{Reference transcript normalization.}
A useful extension would be to develop multi-reference or
normalization-aware evaluation protocols that explicitly account for
transliteration variability in human annotations. Such protocols would
better distinguish genuine recognition errors from acceptable script
or spelling alternatives.

\paragraph{Longitudinal commercial evaluation.}
Because commercial ASR systems change over time, future benchmark
runs should repeat this evaluation across model versions and release
cycles. Longitudinal tracking would make it possible to measure
whether provider updates improve code-switching robustness or simply
shift errors across language pairs and domains.

% =================================================================
\section{Conclusion}
\label{sec:conclusion}
% =================================================================

We have presented a benchmark for evaluating commercial ASR systems
on code-switching speech, covering four language pairs spanning
three scripts and two linguistic families. Our two-stage selection
pipeline selects structurally and semantically difficult samples
while reducing LLM scoring costs by $\approx 91\%$.

ElevenLabs Scribe v2 is the strongest performer across all four
language pairs by both WER and BERTScore, with the gap largest on
Arabic pairs (3.5$\times$ advantage on Egyptian Arabic WER of
13.1\% vs.\ 45.9\% for the next competitor) and smallest on German,
where all but one system achieves under 10\% WER. Azure AI Speech
achieved a mean WER of 43.6\% across all four language pairs after
explicitly enabling \texttt{SpeechService
Connection\_LanguageIdMode=Continuous}
in the SDK configuration --- an improvement of 2.0 percentage points
over the default AtStart LID mode, with the largest gain on
German--English (17.1\% vs.\ 20.5\%).

We demonstrate that WER systematically underestimates the quality
of Arabic and Persian transcriptions due to script convention
variance, and that BERTScore with a multilingual encoder provides
a more reliable quality signal. Crucially, while WER and BERTScore
agree on the ordinal ranking of systems for all Arabic and Persian
pairs ($\tau = 1.0$), WER inflates the magnitude of quality gaps
by a factor of approximately 3$\times$ compared to BERTScore ---
a distinction that matters significantly when selecting between
providers whose absolute performance differences are smaller than
WER suggests.

% =================================================================
\printbibliography

@book{myers1997duelling,
  author    = {Myers-Scotton, Carol},
  title     = {Duelling Languages: Grammatical Structure in Codeswitching},
  year      = {1997},
  publisher = {Oxford University Press},
  address   = {Oxford}
}

@book{auer2013code,
  editor    = {Auer, Peter},
  title     = {Code-Switching in Conversation: Language, Interaction and Identity},
  year      = {2013},
  publisher = {Routledge},
  address   = {London}
}

@book{gumperz1982discourse,
  author    = {Gumperz, John J.},
  title     = {Discourse Strategies},
  year      = {1982},
  publisher = {Cambridge University Press},
  address   = {Cambridge}
}

@article{poplack1980sometimes,
  author  = {Poplack, Shana},
  title   = {Sometimes {I'll} Start a Sentence in {Spanish} {Y} {TERMINO} {EN} {ESPAÑOL}:
             Toward a Typology of Code-Switching},
  journal = {Linguistics},
  volume  = {18},
  number  = {7--8},
  pages   = {581--618},
  year    = {1980}
}

@inproceedings{shia2004language,
  title={Language boundary detection and identification of mixed-language speech based on map estimation},
  author={Shia, Chi-Jiun and Chiu, Yu-Hsien and Hsieh, Jia-Hsin and Wu, Chung-Hsien},
  booktitle={2004 IEEE International Conference on Acoustics, Speech, and Signal Processing},
  volume={1},
  pages={I--381},
  year={2004},
  organization={IEEE}
}

@inproceedings{watanabe2017hybrid,
  author    = {Watanabe, Shinji and Hori, Takaaki and Kim, Suyoun and Hershey, John R. and Hayashi, Tomoki},
  title     = {Hybrid {CTC}/Attention Architecture for End-to-End Speech Recognition},
  booktitle = {Proceedings of Interspeech},
  year      = {2017},
  pages     = {3408--3412}
}

@inproceedings{toshniwal2018multilingual,
  author    = {Toshniwal, Shubham and Sainath, Tara N. and Weiss, Ron J. and Li, Bo and Moreno, Pedro and Weinstein, Eugene and Rao, Kanishka},
  title     = {Multilingual Speech Recognition with a Single End-to-End Model},
  booktitle = {Proceedings of ICASSP},
  year      = {2018},
  pages     = {4904--4908}
}

@article{xie2026switchlingua,
  title   = {SwitchLingua: The First Large-Scale Multilingual and Multi-Ethnic Code-Switching Dataset},
  author  = {Xie, Peng and Liu, Xingyuan and Bie, Yequan and Chan, Tsz Wai and Song, Yangqiu and Wang, Yang and Chen, Hao and Chen, Kani},
  journal = {Advances in Neural Information Processing Systems},
  volume  = {38},
  year    = {2026}
}

@inproceedings{lyu2010seame,
  author    = {Lyu, Dau-Cheng and Tan, Tien-Ping and Chng, Eng Siong and Li, Haizhou},
  title     = {{SEAME}: A {Mandarin-English} Code-Switching Speech Corpus in {South-East Asia}},
  booktitle = {Proceedings of Interspeech},
  year      = {2010},
  pages     = {1986--1989}
}

@misc{deuchar2014bangor,
  author = {Deuchar, Margaret and Davies, Peredur and Herring, Jon and Parafita Couto, Maria Carmen and Carter, Diana},
  title  = {Building Bilingual Corpora},
  year   = {2014},
  note   = {Miami Bangor Corpus}
}

@inproceedings{diwan2021multilingual,
  author    = {Diwan, Anuj and others},
  title     = {Multilingual and Code-Switching {ASR} Challenges for Low Resource {Indian} Languages},
  booktitle = {Proceedings of Interspeech},
  year      = {2021},
  pages     = {2446--2450}
}

@inproceedings{hamed2017collection,
  author    = {Hamed, Injy and Elmahdy, Mohamed and Abdennadher, Slim},
  title     = {Collection and Annotation of {Egyptian Arabic-English} Code-Switched Data},
  booktitle = {Proceedings of the Arabic Natural Language Processing Workshop},
  year      = {2017},
  pages     = {124--130}
}

@inproceedings{hamed-etal-2020-arzen,
    title = "{A}rz{E}n: A Speech Corpus for Code-switched {E}gyptian {A}rabic-{E}nglish",
    author = "Hamed, Injy  and
      Vu, Ngoc Thang  and
      Abdennadher, Slim",
    editor = "Calzolari, Nicoletta  and
      B{\'e}chet, Fr{\'e}d{\'e}ric  and
      Blache, Philippe  and
      Choukri, Khalid  and
      Cieri, Christopher  and
      Declerck, Thierry  and
      Goggi, Sara  and
      Isahara, Hitoshi  and
      Maegaard, Bente  and
      Mariani, Joseph  and
      Mazo, H{\'e}l{\`e}ne  and
      Moreno, Asuncion  and
      Odijk, Jan  and
      Piperidis, Stelios",
    booktitle = "Proceedings of the Twelfth Language Resources and Evaluation Conference",
    month = may,
    year = "2020",
    address = "Marseille, France",
    publisher = "European Language Resources Association",
    url = "https://aclanthology.org/2020.lrec-1.523/",
    pages = "4237--4246",
    language = "eng",
    ISBN = "979-10-95546-34-4"
}

@inproceedings{makhoul1999performance,
  author    = {Makhoul, John and Kubala, Francis and Schwartz, Richard and Weischedel, Ralph},
  title     = {Performance Measures for Information Extraction},
  booktitle = {Proceedings of DARPA Broadcast News Workshop},
  year      = {1999},
  pages     = {249--252}
}

@article{levenshtein1966binary,
  author  = {Levenshtein, Vladimir I.},
  title   = {Binary Codes Capable of Correcting Deletions, Insertions, and Reversals},
  journal = {Soviet Physics Doklady},
  volume  = {10},
  number  = {8},
  pages   = {707--710},
  year    = {1966}
}

@article{zhang2019bertscore,
  title={Bertscore: Evaluating text generation with bert},
  author={Zhang, Tianyi and Kishore, Varsha and Wu, Felix and Weinberger, Kilian Q and Artzi, Yoav},
  journal={arXiv preprint arXiv:1904.09675},
  year={2019}
}

@inproceedings{devlin2019bert,
  author    = {Devlin, Jacob and Chang, Ming-Wei and Lee, Kenton and Toutanova, Kristina},
  title     = {{BERT}: Pre-Training of Deep Bidirectional Transformers for Language Understanding},
  booktitle = {Proceedings of NAACL-HLT},
  year      = {2019},
  pages     = {4171--4186}
}

@article{radford2023robust,
  author  = {Radford, Alec and Kim, Jong Wook and Xu, Tao and Brockman, Greg and McLeavey, Christine and Sutskever, Ilya},
  title   = {Robust Speech Recognition via Large-Scale Weak Supervision},
  journal = {Proceedings of ICML},
  year    = {2023}
}

@inproceedings{baevski2020wav2vec,
  author    = {Baevski, Alexei and Zhou, Yuhao and Mohamed, Abdelrahman and Auli, Michael},
  title     = {wav2vec 2.0: A Framework for Self-Supervised Learning of Speech Representations},
  booktitle = {Advances in Neural Information Processing Systems},
  year      = {2020},
  volume    = {33},
  pages     = {12449--12460}
}

@inproceedings{reimers2019sentencebert,
  author    = {Reimers, Nils and Gurevych, Iryna},
  title     = {Sentence-{BERT}: Sentence Embeddings using {Siamese BERT-Networks}},
  booktitle = {Proceedings of EMNLP-IJCNLP},
  year      = {2019},
  pages     = {3982--3992}
}

@article{mcinnes2018umap,
  author  = {McInnes, Leland and Healy, John and Melville, James},
  title   = {{UMAP}: Uniform Manifold Approximation and Projection for Dimension Reduction},
  journal = {arXiv preprint arXiv:1802.03426},
  year    = {2018}
}

@misc{microsoft2024azurelid,
  author       = {{Microsoft Corporation}},
  title        = {Implement language identification --- {Azure AI Speech}},
  year         = {2024},
  howpublished = {Microsoft Azure Documentation},
  url          = {https://learn.microsoft.com/en-us/azure/ai-services/speech-service/language-identification},
  note         = {Accessed: May 2026}
}

@misc{deepgram2025nova3multi,
  author       = {{Deepgram}},
  title        = {Nova-3 Multilingual Goes {GA}: Real-Time Code-Switching
                  Across 10 Languages},
  year         = {2025},
  howpublished = {Deepgram Developer Changelog},
  url          = {https://developers.deepgram.com/changelog/2025/3/31},
  note         = {Accessed: May 2026}
}

@misc{deepgram2025nova3,
  author       = {{Deepgram}},
  title        = {Introducing {Nova-3}: Setting a New Standard for
                  {AI}-Driven Speech-to-Text},
  year         = {2025},
  howpublished = {Deepgram Blog},
  url          = {https://deepgram.com/learn/introducing-nova-3-speech-to-text-api},
  note         = {Accessed: May 2026}
}

@misc{openai2025transcribe,
  author       = {{OpenAI}},
  title        = {Introducing Next-Generation Audio Models in the {API}},
  year         = {2025},
  howpublished = {OpenAI Blog},
  url          = {https://openai.com/index/introducing-our-next-generation-audio-models/},
  note         = {Accessed: May 2026}
}

@misc{google2025chirp3,
  author       = {{Google Cloud}},
  title        = {Chirp 3 Transcription: Enhanced Multilingual Accuracy},
  year         = {2025},
  howpublished = {Google Cloud Speech-to-Text Documentation},
  url          = {https://docs.cloud.google.com/speech-to-text/docs/models/chirp-3},
  note         = {Accessed: May 2026}
}

@misc{elevenlabs2025scribev2,
  author       = {{ElevenLabs}},
  title        = {Models: {Scribe v2}},
  year         = {2025},
  howpublished = {ElevenLabs Documentation},
  url          = {https://elevenlabs.io/docs/overview/models#scribe-v2},
  note         = {Accessed: May 2026}
}
% =================================================================

% =================================================================
\appendix

\section{ASR Systems and API Parameters}
\label{app:systems_api}

Tables~\ref{tab:systems} and~\ref{tab:api_params} document the
systems evaluated and the exact API parameters used for the benchmark.

\begin{table}[ht]
\centering
\caption{ASR systems evaluated, their underlying architecture,
  and code-switching (CS) language pair support.}
\label{tab:systems}
\begin{tabular}{p{2.0cm}p{2.8cm}p{4.2cm}p{2.2cm}}
\toprule
\textbf{Provider} & \textbf{Model} & \textbf{CS Mechanism} &
  \textbf{CS Pairs} \\
\midrule
ElevenLabs
  & Scribe v2
  & End-to-end multilingual encoder; automatic language detection
    without explicit language input; native intra-utterance
    code-switching across 99 languages
  & All 4 \\[4pt]
OpenAI
  & gpt-4o-transcribe
  & GPT-4o decoder directly integrated into the audio pipeline;
    encoder and language model share weights, enabling contextual
    disambiguation of mixed-language tokens
  & All 4 \\[4pt]
Google
  & Chirp 3
  & Universal Speech Model (USM) trained on 12M hours across 300+
    languages; language-agnostic transcription via
    \apicode{language\_codes=["auto"]} in the Speech-to-Text v2 API
  & All 4 \\[4pt]
Microsoft
  & Azure Speech (CLID)
  & Continuous Language Identification mode
    (\apicode{LanguageIdMode=Continuous}); segments audio at
    utterance level and re-identifies language per segment; up to
    10 candidate languages supplied to \apicode{AutoDetectSourceLanguageConfig}
  & All 4 \\[4pt]
Deepgram
  & Nova-3
  & Unified multilingual decoder with
    \apicode{language=multi}; trained on
    synthetic code-switched data across 10 languages including
    English, Spanish, French, German, Hindi, Russian, Portuguese,
    Japanese, Italian, and Dutch; Arabic and Persian not in the
    documented CS language list
  & German only \\
\bottomrule
\end{tabular}
\end{table}

\begin{table}[ht]
\centering
\caption{API parameters used for each system in this benchmark.
  Parameters not listed were left at provider defaults.
  All audio was resampled to 16\,kHz mono WAV before submission
  to Azure (SDK requirement); other providers received the
  original MP3 files.}
\label{tab:api_params}
\footnotesize
\setlength{\tabcolsep}{4pt}
\begin{tabularx}{\linewidth}{>{\raggedright\arraybackslash}p{2.0cm}>{\raggedright\arraybackslash}p{3.0cm}>{\raggedright\arraybackslash}X}
\toprule
\textbf{Provider} & \textbf{Endpoint / SDK} & \textbf{Key parameters} \\
\midrule
ElevenLabs
  & \apicode{POST}\newline
    \apicode{/v1/speech-to-text}
  & \apicode{model\_id=scribe\_v2}; no \apicode{language\_code} supplied
    (automatic detection); \apicode{diarize=false};
    \apicode{no\_verbatim=false} \\[3pt]
OpenAI
  & \apicode{POST}\newline
    \apicode{/v1/audio/}\newline
    \apicode{/transcriptions}
  & \apicode{model=gpt-4o-transcribe}; no \apicode{language} supplied
    (automatic detection); \apicode{response\_format=text};
    \apicode{temperature=0} \\[3pt]
Google
  & Speech-to-Text v2\newline
    \apicode{Recognize}
  & \apicode{model=chirp\_3}; \apicode{language\_codes=["auto"]}
    (language-agnostic mode); \apicode{auto\_decoding\_config=\{\}};
    region endpoint \apicode{us-speech.googleapis.com} \\[3pt]
Microsoft
  & Azure Speech SDK v1.x\newline
    (Python)
  & \apicode{SpeechServiceConnection\_}\apicode{LanguageIdMode=Continuous}
    (explicitly set via \apicode{set\_property()}; default is \apicode{AtStart});
    \apicode{AutoDetectSource-}\apicode{LanguageConfig} with
    4 candidate locales per dataset
    (\textit{e.g.}, \apicode{ar-EG, en-US} for Egyptian Arabic);
    continuous recognition; workers forced to 1 to avoid
    macOS SDK audio concurrency issue \cite{microsoft2024azurelid} \\[3pt]
Deepgram
  & \apicode{POST}\newline
    \apicode{/v1/listen}
  & \apicode{model=nova-3}; \apicode{language=multi}
    (multilingual code-switching mode); \apicode{smart\_format=true};
    \apicode{punctuate=true} \\
\bottomrule
\end{tabularx}
\end{table}

% =================================================================

\section{H\_Score Formula and Signal Weights}
\label{app:hscore}

Full H\_Score composite (Equation~\ref{eq:hscore}):

\begin{align*}
  H &= 0.30\,h_\text{mix}
     + 0.30\,h_\text{alt}
     + 0.20\,h_\text{morph}
     + 0.10\,h_\text{len}
     + 0.10\,h_\text{vocab} \\[6pt]
  h_\text{mix}   &= \min\!\left(\tfrac{\min(n_a,n_l)/(n_a+n_l)}{0.35},\,1\right)\times 10 \\
  h_\text{alt}   &= \min\!\left(\tfrac{k}{n/2},\,1\right)\times 10 \\
  h_\text{morph} &= \min\!\left(\tfrac{b}{3},\,1\right)\times 10 \\
  h_\text{len}   &= \begin{cases}0 & n < 5 \\
                    \min\!\left(\tfrac{n-5}{20},1\right)\times 10 & n \ge 5
                    \end{cases} \\
  h_\text{vocab} &= \min\!\left(\tfrac{|\text{vocab}|/n}{0.7},\,1\right)\times 10
\end{align*}

\section{LLM Scoring Prompt Structure}
\label{app:prompt}

Each LLM receives the transcript text and language pair label and
returns a JSON response with the following schema (temperature 0.1):

\begin{lstlisting}[language={}]
{
  "overall_score": <int 1-10>,
  "dimensions": {
    "morphological_blending":         {"score":<1-10>, "evidence":"..."},
    "switching_density":              {"score":<1-10>, "evidence":"..."},
    "slang_and_register_mix":         {"score":<1-10>, "evidence":"..."},
    "phonological_ambiguity":         {"score":<1-10>, "evidence":"..."},
    "named_entity_jargon_density":    {"score":<1-10>, "evidence":"..."},
    "script_orthographic_complexity": {"score":<1-10>, "evidence":"..."}
  },
  "hard_tokens": [{"token":"...", "reason":"..."}, ...],
  "summary": "..."
}
\end{lstlisting}

The Ensemble\_Score is the average of both \texttt{overall\_score}
values. Rows where the two models disagree by $>3$ points on any
dimension are flagged for manual review.

\section{Qualitative Transcription Comparison}
\label{app:qualitative}

Table~\ref{tab:qualitative_full} presents a comprehensive side-by-side
comparison across \textbf{all four language pairs} and \textbf{all
CS-supported models}. For each language pair, the 5 samples with the
highest WER--BERTScore divergence are selected --- these are
the cases where WER most severely over-penalises a semantically
correct transcription due to transliteration variance.

The $\Delta$ column is the key signal, defined as:
\begin{equation}
  \Delta = \text{WER} - (1 - F_\text{BERT})
  \label{eq:delta}
\end{equation}
where $1 - F_\text{BERT}$ is the \emph{semantic error rate} implied
by BERTScore --- converting it to the same scale as WER (lower = better).
Interpreting $\Delta$:
\begin{itemize}[leftmargin=*]
  \item $\Delta > 0$: WER penalises the transcription \emph{more} than
        its semantic distance from the reference justifies.
        This is the signature of transliteration variance --- the model
        produced the correct meaning in a different script convention.
  \item $\Delta \approx 0$: both metrics agree; the transcription
        error is genuine, not a script-convention artefact.
  \item $\Delta < 0$: BERTScore detects more semantic distance
        than WER counts as word errors (rare).
\end{itemize}
Rows with $\Delta > 0.10$ are shown in \textbf{bold}.
Deepgram is included for German only (no Arabic/Persian CS support).
\begin{landscape}
\small
\setlength{\LTcapwidth}{\linewidth}
\begin{longtable}{p{3.5cm}p{2.0cm}p{3.5cm}p{1.0cm}p{1.0cm}p{1.1cm}}
\caption{Comprehensive qualitative comparison across all four language pairs and CS-supported models. Rows are selected to maximise WER--BERTScore divergence ($\Delta = \text{WER} - (1 - F_\text{BERT})$): a positive $\Delta$ means WER penalises the transcription more than its semantic error rate justifies --- the signature of transliteration variance. Bold = $\Delta > 0.10$. Deepgram shown for German only.}\\
\label{tab:qualitative_full}\\
\toprule
\textbf{Reference} & \textbf{Model} & \textbf{Hypothesis} & \textbf{WER} & \textbf{BERT} & \textbf{$\Delta$} \\
\midrule
\endfirsthead
\multicolumn{6}{c}{\tablename~\thetable{} -- continued}\\
\toprule
\textbf{Reference} & \textbf{Model} & \textbf{Hypothesis} & \textbf{WER} & \textbf{BERT} & \textbf{$\Delta$} \\
\midrule
\endhead
\midrule \multicolumn{6}{r}{Continued on next page}\\
\endfoot
\bottomrule
\endlastfoot
\multicolumn{6}{l}{\textbf{Egyptian Arabic--English}\hspace{0.5em}\textit{(5 samples with highest WER--BERTScore divergence)}}\\[2pt]
\midrule
\rtlcell{في \LR{bug} غريب في السيستم بيعمل \LR{logout} للـ \LR{users}، لازم نعمله \LR{fix} في أسرع وقت.} & {\small ElevenLabs} & \rtlcell{في \LR{bug} غريب في الـ \LR{system} بيعمل \LR{log out} للـ \LR{users}. لازم نعمل له \LR{fix} في أسرع وقت} & {\small \textbf{46.7\%}} & {\small \textbf{0.879}} & {\small \textbf{+0.346}} \\
 & {\small OpenAI} & \rtlcell{في بگ غريب في السيستم بيعمل لوج اوت لليوزرز. لازم نعمله فيكس في أسرع وقت.} & {\small \textbf{33.3\%}} & {\small \textbf{0.832}} & {\small \textbf{+0.165}} \\
 & {\small Google} & \rtlcell{في \LR{bug} غريب في السيستم بيعمل \LR{log out} لليوزرز، لازم نعمل له \LR{fix} في أسرع وقت.} & {\small \textbf{33.3\%}} & {\small \textbf{0.927}} & {\small \textbf{+0.260}} \\
 & {\small Azure} & \rtlcell{في باج غريب في السيستم بيعمل لوج أوت لليوزر؟ لازم نعمله في أسرع وقت؟} & {\small \textbf{40.0\%}} & {\small \textbf{0.785}} & {\small \textbf{+0.185}} \\
[2pt]
\rtlcell{الـ \LR{internet connection} وحش جداً النهاردة، حاولت أعمل \LR{troubleshooting} للراوتر بس مفيش فايدة.} & {\small ElevenLabs} & \rtlcell{الـ \LR{internet connection} وحش جدا النهاردة. حاولت أعمل \LR{troubleshooting} للـ \LR{router} بس ما فيش فايدة} & {\small \textbf{46.2\%}} & {\small \textbf{0.884}} & {\small \textbf{+0.345}} \\
 & {\small OpenAI} & \rtlcell{الإنترنت كونكشن وحش جدا النهاردة. حاولت أعمل تربل شوتينج للراوتر بس مفيش فايدة.} & {\small \textbf{53.8\%}} & {\small \textbf{0.875}} & {\small \textbf{+0.414}} \\
 & {\small Google} & \rtlcell{الانترنت كونكشن وحش جدا النهارده حاولت اعمل ترابل شوتنج للراوتر بس ما فيش فائده} & {\small \textbf{84.6\%}} & {\small \textbf{0.773}} & {\small \textbf{+0.619}} \\
 & {\small Azure} & \rtlcell{الإنترنت كونكشن وحش جد ا النهارده؟ حاولت أعمل ترابل شوتنج للراوتر بس ما فيش فايدة.} & {\small \textbf{76.9\%}} & {\small \textbf{0.826}} & {\small \textbf{+0.595}} \\
[2pt]
\rtlcell{الـ \LR{new feature} اللي بنعملها دي هتحتاج \LR{integration} مع \LR{API} خارجي.} & {\small ElevenLabs} & \rtlcell{هني يفيد شو اللي بنعملها؟ إذا بدك \LR{integration} مع \LR{API} خارجي} & {\small \textbf{54.5\%}} & {\small \textbf{0.775}} & {\small \textbf{+0.321}} \\
 & {\small OpenAI} & \rtlcell{اللي بتعملها ده تحتاج انتجريشن مع \LR{API} خارجي} & {\small \textbf{63.6\%}} & {\small \textbf{0.822}} & {\small \textbf{+0.459}} \\
 & {\small Google} & \rtlcell{\LR{new feature}} & {\small \textbf{81.8\%}} & {\small \textbf{0.733}} & {\small \textbf{+0.551}} \\
 & {\small Azure} & \rtlcell{أن يفيد شغل ابن منازل، تحتك، \LR{integration}، \LR{my} خارجي.} & {\small \textbf{90.9\%}} & {\small \textbf{0.704}} & {\small \textbf{+0.613}} \\
[2pt]
\rtlcell{لازم نعمل \LR{commit} للكود ده على الـ \LR{main branch} قبل الـ \LR{deadline} بتاع بكرة.} & {\small ElevenLabs} & \rtlcell{لازم نعمل \LR{commit} لـ \LR{code} ده على \LR{Main branch} قبل \LR{deadline} بتاعكوا.} & {\small \textbf{42.9\%}} & {\small \textbf{0.880}} & {\small \textbf{+0.308}} \\
 & {\small OpenAI} & \rtlcell{لازم نعمل \LR{commit} للكود ده للمين برانش قبل الديدلاين بتاع بكره.} & {\small \textbf{50.0\%}} & {\small \textbf{0.847}} & {\small \textbf{+0.347}} \\
 & {\small Google} & \rtlcell{لازم نعمل كوميت للكود ده للمين برانش قبل الديد لاين بتاع بكره.} & {\small \textbf{57.1\%}} & {\small \textbf{0.792}} & {\small \textbf{+0.363}} \\
 & {\small Azure} & \rtlcell{\LR{Azam Neville}, \LR{Comet}, \LR{Liquid}, \LR{Dallas}, \LR{main branch of the deadline with ABUK}.} & {\small \textbf{78.6\%}} & {\small \textbf{0.678}} & {\small \textbf{+0.464}} \\
[2pt]
\rtlcell{أنا حاسس إن الـ \LR{account} بتاعي اتعمله \LR{hacked} ومش عارف أعمل \LR{log in}.} & {\small ElevenLabs} & \rtlcell{أنا حاسس إن الأكونت بتاعي اتعمل له \LR{hack} ومش عارف أعمل \LR{log in}.} & {\small \textbf{38.5\%}} & {\small \textbf{0.922}} & {\small \textbf{+0.307}} \\
 & {\small OpenAI} & \rtlcell{أنا حاسس إن الأكاونت بتاعي اتعمله هاكد ومش عارف أعمل لوج إن.} & {\small \textbf{38.5\%}} & {\small \textbf{0.827}} & {\small \textbf{+0.211}} \\
 & {\small Google} & \rtlcell{انا حاسس ان الاكونت بتاعي اتعمل له هاجد ومش عارف اعمل لوجن.} & {\small \textbf{76.9\%}} & {\small \textbf{0.814}} & {\small \textbf{+0.583}} \\
 & {\small Azure} & \rtlcell{أنا حاسس أنه الأكاونت بتاعي. أتعمله هاكد ومش عارف أعمل لوج إن.} & {\small \textbf{53.8\%}} & {\small \textbf{0.801}} & {\small \textbf{+0.340}} \\
[2pt]
\midrule[0.8pt]
\multicolumn{6}{l}{\textbf{Saudi Arabic--English}\hspace{0.5em}\textit{(5 samples with highest WER--BERTScore divergence)}}\\[2pt]
\midrule
\rtlcell{أكثر شي يتعبني يوم يصير \LR{bug} غريب، أجلس أسوي \LR{check} للـ \LR{logs} وأرجع للـ \LR{commit} القديم، وإذا انحلت المشكلة أحس \LR{satisfaction} عالي كأني جبت \LR{achievement}.} & {\small ElevenLabs} & \rtlcell{أكثر شيء يتعبني ما يصير \LR{bug} غريب. أجلس أسوي \LR{check} لل\LR{ogs} وأرجع لل\LR{commit} القديم إذا حلت المشكلة حيس ما تسواكش العلي كأنك تبدأ \LR{treatment}} & {\small \textbf{68.0\%}} & {\small \textbf{0.838}} & {\small \textbf{+0.518}} \\
 & {\small OpenAI} & \rtlcell{أكثر شيء يتعبني من صير فترة غريب أجلس أسوي تاك للوغس وأرجع للكومنتات القديمة إذا حليت المشكلة حسة سواتشن علي كأن تيب تا تيمنت.} & {\small \textbf{80.0\%}} & {\small \textbf{0.766}} & {\small \textbf{+0.566}} \\
 & {\small Google} & \rtlcell{اكثر شيء يتعبني يوم يصير بق غريب اجلس اسوي تشيك لللوكس وارجع للكوميت القديمه اذا انحلت المشكله احس الساتيسفاكشن عالي كاني جبت اتشييفمنت.} & {\small \textbf{76.0\%}} & {\small \textbf{0.754}} & {\small \textbf{+0.514}} \\
 & {\small Azure} & \rtlcell{أكثر شيء يتعبني، ويصيرك غريب، أجلس، أسوي تشيك لللوكس، وأرجع للكوميدي القديم إلى حالة المشكلة، أحسها ديسفاكشن علي كأني تبدأ تشيف ميند.} & {\small \textbf{84.0\%}} & {\small \textbf{0.770}} & {\small \textbf{+0.610}} \\
[2pt]
\rtlcell{إذا البلوتوث ما يلقط السماعة أسوي \LR{forget device} وأعيد \LR{pairing}، وأتأكد من \LR{battery}، وإذا استمرت المشكلة أجرب \LR{firmware update} عشان يرجع \LR{stable}.} & {\small ElevenLabs} & \rtlcell{إذا الـ \LR{Bluetooth} ما يلد السماعة، سوي فرقع \LR{Device} وأعيد \LR{pairing} وتأكد من \LR{Battery}. وإذا استمرت المشكلة، جرب \LR{Firmware Update} عشان يجي أستبل.} & {\small \textbf{59.1\%}} & {\small \textbf{0.836}} & {\small \textbf{+0.427}} \\
 & {\small OpenAI} & \rtlcell{إذا البلوتوث ما يلقط السماعة سوي فرقت ديفايس وأعيد بيرينغ وتأكد من البطاري وإذا استمرت المشكلة جرب فيرموير أبديت عشان يرجع ستايبل.} & {\small \textbf{45.5\%}} & {\small \textbf{0.827}} & {\small \textbf{+0.281}} \\
 & {\small Google} & \rtlcell{اذا البلوتوث ما لقط السماعه اسوي فورجيت ديفايس واعيد بيرينج واتاكد من باتري، اذا استمرت المشكله اجرب فيرم وير ابديت عشان يرجع ستيب.} & {\small \textbf{77.3\%}} & {\small \textbf{0.761}} & {\small \textbf{+0.534}} \\
 & {\small Azure} & \rtlcell{إذا البلوتوث ما يلقط السماعة أسوي فرق الدفايس، وأعيد بيرنج، وأتأكد من بتار، إذا استمرت المشكلة. جرب فيرميور وإبديت عشان يرجع سيب.} & {\small \textbf{40.9\%}} & {\small \textbf{0.833}} & {\small \textbf{+0.242}} \\
[2pt]
\rtlcell{النت إذا صار \LR{slow} وأنا عندي تسليم، أول شي أسوي \LR{troubleshooting}: أفصل الراوتر وأرجعه، وأشيك \LR{settings}، وإذا ما نفع أحول \LR{hotspot} كـ \LR{quick fix}.} & {\small ElevenLabs} & \rtlcell{النت إذا صار \LR{slow} عند تسليم أول شيء اسوي \LR{troubleshooting}. افصل \LR{Router} ارده معه \LR{Check settings} وإذا ما نفع حاول \LR{hotspot} ك \LR{Quick fix}.} & {\small \textbf{54.2\%}} & {\small \textbf{0.834}} & {\small \textbf{+0.375}} \\
 & {\small OpenAI} & \rtlcell{النت إذا صار \LR{slow} عندك أول شيء سوي \LR{troubleshooting} افصل الراوتر وارجع عشيك \LR{settings} وإذا ما نفع حاول \LR{hotspot} ك \LR{quick fix}.} & {\small \textbf{45.8\%}} & {\small \textbf{0.860}} & {\small \textbf{+0.318}} \\
 & {\small Google} & \rtlcell{النت اذا صار سلو عندي تسليم اول شيء اسوي تروبل شوتنج افصل الراوتر وارجعه واشيك سيتنج واذا ما نفع احاول هوت سبوت ككويك فيكس.} & {\small \textbf{79.2\%}} & {\small \textbf{0.744}} & {\small \textbf{+0.536}} \\
 & {\small Azure} & \rtlcell{النت إذا صار سلون عند تسليم أول شيء تسوي التروب للشوتين، افصل الراوتر ورجع، وإذا ما نفع حاول هوتسبوك ويكفيكس.} & {\small \textbf{70.8\%}} & {\small \textbf{0.794}} & {\small \textbf{+0.502}} \\
[2pt]
\rtlcell{التطبيق يطلع \LR{error} كل ما أحاول أسوي \LR{login}، جربت \LR{password} جديد وما ضبط، قلت يمكن \LR{server} فيه ضغط، فانتظرت شوي وسويت \LR{refresh}، وبالأخير احتجت \LR{verification} مرة ثانية.} & {\small ElevenLabs} & \rtlcell{التطبيق ضل عير الكلمة عسوي \LR{Login}. جربت \LR{password} جديد وما عجبك. قلت يمكن \LR{server} فيه ضغط فانتظرت شوية وسويت \LR{refresh} وباقي راح تجد \LR{verification} مرة ثانية.} & {\small \textbf{51.8\%}} & {\small \textbf{0.854}} & {\small \textbf{+0.372}} \\
 & {\small OpenAI} & \rtlcell{المشكلة لما احاول اسوي لوق ان جربت باسورد جديد وما ضبط قلت يمكن السيرفر فيه ضغط فانتظرت شوي وسويت ريفريش وبالاخير احتجت فيريفكيشن مرة ثانية} & {\small \textbf{55.6\%}} & {\small \textbf{0.779}} & {\small \textbf{+0.335}} \\
 & {\small Google} & \rtlcell{تطبيق يطلع ايرور كل ما اسوي لوجن جربت باسورد جديد وما ضبط قلت يمكن سيرفر في ضغط فانتظرت شوي وسويت ريفريش وبالاخير احتاج فيرفكيشن مره ثانيه.} & {\small \textbf{59.3\%}} & {\small \textbf{0.784}} & {\small \textbf{+0.377}} \\
 & {\small Azure} & \rtlcell{تطبيق طلع عروض كل ما أسوي لوجن. جربت باسوورد جديد وما ظبط قد يمكن في ضغط. فانتظرت شوية وسويت ريفريش. وبالأخير احتجت مرة ثانيه.} & {\small \textbf{55.6\%}} & {\small \textbf{0.807}} & {\small \textbf{+0.363}} \\
[2pt]
\rtlcell{قبل ما أبدأ لعب أشيك \LR{ping} وأضبط \LR{loadout} على حسب الـ \LR{meta}، مو لازم \LR{rush} دايم، الأهم \LR{strategy} واضحة و\LR{team work} مضبوط عشان نجيب \LR{win} نظيف.} & {\small ElevenLabs} & \rtlcell{قبل ما أضبط لعبة أشيك \LR{ping} وأضبط \LR{load out} على حسب المפה، ولازم \LR{rush} دايما، أهم \LR{strategy} واضحة و\LR{teamwork} مضبوط عشان نجيب \LR{win} نظيف.} & {\small \textbf{46.2\%}} & {\small \textbf{0.888}} & {\small \textbf{+0.350}} \\
 & {\small OpenAI} & \rtlcell{قبل ما اضبط اللعبة شيك بنق واضبط لوداوت على حسب المتا ولازم رش دايما اهم استراتيجية واضحة وتيم وورك مضبوط عشان نجيب ون نظيف} & {\small \textbf{65.4\%}} & {\small \textbf{0.744}} & {\small \textbf{+0.398}} \\
 & {\small Google} & \rtlcell{قبل ما اضبط لعبه اشيك بنج واضبط لود اوت على حسب المتا، مو لازم رش دائما اهم استراتيجي واضحه وتيم ورك مضبوط عشان نجيب وين نظيف.} & {\small \textbf{65.4\%}} & {\small \textbf{0.754}} & {\small \textbf{+0.408}} \\
 & {\small Azure} & \rtlcell{قبل ما اضبط لعبة شيك بنجو اضبط لودعوت على حسب المتى ولازم نعيش دائما أهم استراتيجي واضحة وتيم وورك مظبوط عشان نجيب وين نظيف.} & {\small \textbf{69.2\%}} & {\small \textbf{0.735}} & {\small \textbf{+0.427}} \\
[2pt]
\midrule[0.8pt]
\multicolumn{6}{l}{\textbf{Persian (Farsi)--English}\hspace{0.5em}\textit{(5 samples with highest WER--BERTScore divergence)}}\\[2pt]
\midrule
\rtlcell{لطفا قبل از \LR{meeting} بعدی، روی \LR{feedback} هایی که دادم کار کنید.} & {\small ElevenLabs} & \rtlcell{لطفاً قبل از \LR{meeting} بعدی روی \LR{feedback}هایی که داده‌ام کار کنید} & {\small \textbf{50.0\%}} & {\small \textbf{0.887}} & {\small \textbf{+0.387}} \\
 & {\small OpenAI} & \rtlcell{لوچسان بعض ميگن كه بعدی روی سفید بدهایک دادن کار کنید.} & {\small \textbf{75.0\%}} & {\small \textbf{0.710}} & {\small \textbf{+0.460}} \\
 & {\small Google} & \rtlcell{لطفاً قبل از میدینگ بعدی روی فیدبک‌هایی که دادم کار کنید.} & {\small \textbf{33.3\%}} & {\small \textbf{0.863}} & {\small \textbf{+0.196}} \\
 & {\small Azure} & \rtlcell{لطفا قبل از میتینگ بعدی روی فیدبک هایی که دادم کار کنید.} & {\small \textbf{25.0\%}} & {\small \textbf{0.895}} & {\small \textbf{+0.145}} \\
[2pt]
\rtlcell{اینقدر برای \LR{final exam} استرس دارم که اصلا نمیتونم روی درسام \LR{focus} کنم.} & {\small ElevenLabs} & \rtlcell{اینقدر برای فینال اگزام استرس دارم که اصلا نمی‌تونم روی درس‌هام فوکس کنم} & {\small \textbf{53.8\%}} & {\small \textbf{0.848}} & {\small \textbf{+0.386}} \\
 & {\small OpenAI} & \rtlcell{اینقدر برای فاینال اگزم استرس دارم که اصلا نمی‌تونم روی درسام فوکوس کنم.} & {\small \textbf{38.5\%}} & {\small \textbf{0.883}} & {\small \textbf{+0.267}} \\
 & {\small Google} & \rtlcell{این‌قدر برای فاینال اگزام استرس دارم که اصلاً نمی‌تونم روی درس‌هام فوکوس کنم.} & {\small \textbf{76.9\%}} & {\small \textbf{0.876}} & {\small \textbf{+0.645}} \\
 & {\small Azure} & \rtlcell{اینقدر برای فایل عزیزم استرس دارم که اصلا نمیتونم روی درسام فکس کنم.} & {\small \textbf{23.1\%}} & {\small \textbf{0.901}} & {\small \textbf{+0.132}} \\
[2pt]
\rtlcell{این \LR{character} جدید \LR{skill} های خیلی خفنی داره، باید حتماً \LR{unlock} اش کنم.} & {\small ElevenLabs} & \rtlcell{این کاراکتر جدید \LR{Skill}های خیلی خفنی داره باید حتما \LR{Unlock}ش کنم} & {\small \textbf{53.8\%}} & {\small \textbf{0.847}} & {\small \textbf{+0.385}} \\
 & {\small OpenAI} & \rtlcell{این کاراکتر جدید اسکیل‌های خیلی خفنی داره، باید حتما آنلاکش کنم.} & {\small \textbf{38.5\%}} & {\small \textbf{0.859}} & {\small \textbf{+0.244}} \\
 & {\small Google} & \rtlcell{این کاراکتر جدید اسکیل‌های خیلی خفنی داره باید حتماً آنلاک‌اش کنم.} & {\small \textbf{30.8\%}} & {\small \textbf{0.844}} & {\small \textbf{+0.152}} \\
 & {\small Azure} & \rtlcell{این کرکتر جدید اسکیل های خیلی خفن داره باید حتما لاکش کند.} & {\small \textbf{61.5\%}} & {\small \textbf{0.811}} & {\small \textbf{+0.426}} \\
[2pt]
\rtlcell{این \LR{weekend} میخوام کاملاً \LR{chill} کنم و چندتا \LR{movie} جدید ببینم.} & {\small ElevenLabs} & \rtlcell{این \LR{weekend} می‌خوام کاملاً چیل کنم و چند تا \LR{movie} جدید ببینم} & {\small \textbf{45.5\%}} & {\small \textbf{0.909}} & {\small \textbf{+0.364}} \\
 & {\small OpenAI} & \rtlcell{این ویکند خانواده‌امشیطانند و چندتا مهموی جالب بینند.} & {\small \textbf{72.7\%}} & {\small \textbf{0.706}} & {\small \textbf{+0.434}} \\
 & {\small Google} & \rtlcell{این ویکند می‌خوام کاملاً چیل کنم و چند تا مووی جدید ببینم.} & {\small \textbf{63.6\%}} & {\small \textbf{0.873}} & {\small \textbf{+0.510}} \\
 & {\small Azure} & \rtlcell{این ویکند میخوام کاملا چه کنم چندتا مویی جدید ببینم.} & {\small \textbf{45.5\%}} & {\small \textbf{0.820}} & {\small \textbf{+0.275}} \\
[2pt]
\rtlcell{این \LR{influencer} جدید خیلی \LR{content} جالبی تولید می‌کنه، همه‌ش رو دنبال می‌کنم.} & {\small ElevenLabs} & \rtlcell{این اینفلوئنسر جدید خیلی \LR{content} جالبی تولید میکنه. همش رو دنبال میکنم} & {\small \textbf{46.7\%}} & {\small \textbf{0.857}} & {\small \textbf{+0.324}} \\
 & {\small OpenAI} & \rtlcell{این اینفلوئنسر جدید خیلی کانتنت جالبی تولید می‌کنه. همش رو دنبال می‌کنم.} & {\small \textbf{33.3\%}} & {\small \textbf{0.876}} & {\small \textbf{+0.209}} \\
 & {\small Google} & \rtlcell{این اینفلوئنسر جدید خیلی کانتنت جالبی تولید می‌کنه. همه‌شو دنبال می‌کنم.} & {\small \textbf{33.3\%}} & {\small \textbf{0.871}} & {\small \textbf{+0.205}} \\
 & {\small Azure} & \rtlcell{این پل عنصر جدید خیلی کامنت جالبی تولید میکنه همش را دنبال میکنند.} & {\small \textbf{66.7\%}} & {\small \textbf{0.836}} & {\small \textbf{+0.502}} \\
[2pt]
\midrule[0.8pt]
\multicolumn{6}{l}{\textbf{German--English}\hspace{0.5em}\textit{(5 samples with highest WER--BERTScore divergence)}}\\[2pt]
\midrule
{\small Ich habe nächste Woche einen appointment für einen allgemeinen check-up beim Arzt.} & {\small ElevenLabs} & {\small Ich habe nächste Woche ein Appointment für einen allgemeinen Checkup beim Arzt.} & {\small \textbf{23.1\%}} & {\small \textbf{0.946}} & {\small \textbf{+0.177}} \\
 & {\small OpenAI} & {\small Ich habe nächste Woche ein Appointment für einen allgemeinen Check-up beim Arzt.
} & {\small 7.7\%} & {\small 0.962} & {\small +0.038} \\
 & {\small Google} & {\small Ich habe nächste Woche einen Appointment für einen allgemeinen Check-up beim Arzt.} & {\small 0.0\%} & {\small 0.965} & {\small -0.035} \\
 & {\small Azure} & {\small Ich habe nächste Woche einen appointment für einen allgemeinen Checkup beim Arzt.} & {\small \textbf{15.4\%}} & {\small \textbf{0.949}} & {\small \textbf{+0.102}} \\
 & {\small Deepgram} & {\small Ich habe nächste Woche ein Appointment für einen allgemeinen Check-up beim Arzt.} & {\small 7.7\%} & {\small 0.962} & {\small +0.038} \\
[2pt]
{\small Für das Meeting morgen müssen wir die key takeaways nochmal genau durchgehen.} & {\small ElevenLabs} & {\small Für das Meeting morgen müssen wir die Key Takeaways noch mal genau durchgehen.} & {\small \textbf{16.7\%}} & {\small \textbf{0.975}} & {\small \textbf{+0.142}} \\
 & {\small OpenAI} & {\small Für das Meeting morgen müssen wir die Key Takeaways nochmal genau durchgehen.
} & {\small 0.0\%} & {\small 0.983} & {\small -0.017} \\
 & {\small Google} & {\small Für das Meeting morgen müssen wir die Key Takeaways noch mal genau durchgehen.} & {\small \textbf{16.7\%}} & {\small \textbf{0.975}} & {\small \textbf{+0.142}} \\
 & {\small Azure} & {\small Für das Meeting morgen müssen wir die Key Tech raise noch mal genau durchgehen.} & {\small \textbf{33.3\%}} & {\small \textbf{0.930}} & {\small \textbf{+0.263}} \\
 & {\small Deepgram} & {\small Für das Meeting morgen müssen wir die Key Takeaways nochmal genau durchgehen.} & {\small 0.0\%} & {\small 0.983} & {\small -0.017} \\
[2pt]
{\small My doctor said I need to get a Bluttest to check my iron levels.} & {\small ElevenLabs} & {\small My doctor said I need to get a blood test to check my iron levels.} & {\small \textbf{14.3\%}} & {\small \textbf{0.967}} & {\small \textbf{+0.110}} \\
 & {\small OpenAI} & {\small My doctor said I need to get a blood test to check my iron levels.
} & {\small \textbf{14.3\%}} & {\small \textbf{0.967}} & {\small \textbf{+0.110}} \\
 & {\small Google} & {\small My doctor said I need to get a blood test to check my iron levels.} & {\small \textbf{14.3\%}} & {\small \textbf{0.967}} & {\small \textbf{+0.110}} \\
 & {\small Azure} & {\small My Doctor. I need to get a blood test to check my iron levels.} & {\small \textbf{21.4\%}} & {\small \textbf{0.913}} & {\small \textbf{+0.128}} \\
 & {\small Deepgram} & {\small My doctor said I need to get a blood test to check my iron levels.} & {\small \textbf{14.3\%}} & {\small \textbf{0.967}} & {\small \textbf{+0.110}} \\
[2pt]
{\small Wir haben ein Upgrade auf eine Suite mit king-size bed und tollem sea view bekommen.} & {\small ElevenLabs} & {\small Wir haben ein Upgrade auf eine Suite mit Kingsize-Bett und tollem Sea View bekommen.} & {\small \textbf{18.8\%}} & {\small \textbf{0.917}} & {\small \textbf{+0.105}} \\
 & {\small OpenAI} & {\small Wir haben ein Upgrade auf eine Suite mit Kingsize-Bett und Dorm Sea View bekommen.
} & {\small \textbf{25.0\%}} & {\small \textbf{0.889}} & {\small \textbf{+0.139}} \\
 & {\small Google} & {\small Wir haben ein Upgrade auf eine Suite mit Kingsize-Bett und Dolm Sea View bekommen.} & {\small \textbf{25.0\%}} & {\small \textbf{0.891}} & {\small \textbf{+0.141}} \\
 & {\small Azure} & {\small Wir haben ein Upgrade auf eine Suite mit King Size Bad. Dorm C View become an.} & {\small \textbf{37.5\%}} & {\small \textbf{0.821}} & {\small \textbf{+0.196}} \\
 & {\small Deepgram} & {\small Wir haben ein Upgrade auf eine Suite mit King-size Bett und Dorm Sea View bekommen.} & {\small 12.5\%} & {\small 0.926} & {\small +0.051} \\
[2pt]
{\small We should plan a proper Kaffeeklatsch soon, with lots of cake and extra Sahne.} & {\small ElevenLabs} & {\small We should plan a proper coffee klatch soon, with lots of cake and extra Sahne.} & {\small 14.3\%} & {\small 0.949} & {\small +0.092} \\
 & {\small OpenAI} & {\small We should planen a proper Kaffeeklatsch soon, with lots of cake and extra Sahne.
} & {\small 7.1\%} & {\small 0.980} & {\small +0.051} \\
 & {\small Google} & {\small We should plan a proper coffee klatsch soon with a lot of cake and extra Sahne.} & {\small \textbf{28.6\%}} & {\small \textbf{0.928}} & {\small \textbf{+0.214}} \\
 & {\small Azure} & {\small We should pläne proper kaffee klatscht duon with a lots of cake and extra sahne.} & {\small \textbf{42.9\%}} & {\small \textbf{0.823}} & {\small \textbf{+0.252}} \\
 & {\small Deepgram} & {\small We should plan a proper Kaffeeklatch soon, with lots of cake and extra Sahne.} & {\small 7.1\%} & {\small 0.994} & {\small +0.065} \\
[2pt]
\end{longtable}
\end{landscape}

\end{document}